\documentclass[twoside,11pt]{article}

\usepackage{jmlr2e}
\usepackage{amsmath}
\usepackage{mathtools}
\usepackage{tabularx}
\usepackage[usenames,dvipsnames]{xcolor}
\usepackage{booktabs}
\usepackage{enumerate}
\mathtoolsset{showonlyrefs}
\usepackage[linesnumbered,ruled]{algorithm2e}

\newcommand{\dataset}{{\cal D}}

\newcommand{\EE}[2]{\mathbb{E}_{#1}\left[ #2 \right]}
\newcommand{\set}[1]{\left\lbrace #1 \right\rbrace}
\newcommand{\bs}[1]{\boldsymbol{#1}}
\newcommand{\bx}{\boldsymbol{x}}
\newcommand{\bz}{\boldsymbol{z}}
\newcommand{\bmu}{\boldsymbol{\mu}}
\newcommand{\btheta}{\boldsymbol{\theta}}
\newcommand{\bomega}{\boldsymbol{\omega}}
\newcommand{\bg}{\boldsymbol{g}}
\newcommand{\bnu}{\boldsymbol{\nu}}
\newcommand{\bks}{\boldsymbol{k}_s}
\newcommand{\PG}{p_{\scriptscriptstyle \mathrm{PG}}}

\newcommand{\Po}{p_\mathrm{{\tiny po}}}

\newcommand{\calZ}{\mathcal{Z}}
\newcommand{\R}{\mathbb{R}}
\newcommand{\calH}{\mathcal{H}}
\newcommand{\calM}{\mathcal{M}}
\newcommand{\X}{\mathcal{X}}
\newcommand{\Q}{\mathcal{Q}}

\binoppenalty=\maxdimen
\relpenalty=\maxdimen

\usepackage{lastpage}

\ShortHeadings{Sigmoidal Gaussian Cox Process Inference}{Donner and Opper}
\firstpageno{1}

\begin{document}

\title{Efficient Bayesian Inference of Sigmoidal Gaussian Cox Processes}

\author{\name Christian Donner \email christian.donner@bccn-berlin.de \\
		\name Manfred Opper \email manfred.opper@tu-berlin.de \\
       \addr Artificial Intelligence Group\\
       Technische Universit\"at Berlin\\
       Berlin, Germany}

\editor{}

\maketitle

\begin{abstract}
We present an approximate Bayesian inference approach for estimating the intensity
of an inhomogeneous Poisson process, where the intensity function is modelled using a Gaussian process (GP) prior via a sigmoid link function. Augmenting the model using a latent marked Poisson process and P\'olya--Gamma random variables we obtain a representation of the likelihood which is conjugate to the GP prior. We estimate the posterior using a variational free--form mean field optimisation together with the framework of sparse GPs. Furthermore, as alternative approximation we suggest a sparse Laplace's method for the posterior, for which an efficient expectation--maximisation algorithm is derived to find the posterior's mode. Both algorithms compare well against exact inference obtained by a Markov Chain Monte Carlo sampler and standard variational Gauss approach solving the same model, while being one order of magnitude faster. Furthermore, the performance and speed of our method is competitive with that of another recently proposed Poisson process model based on a quadratic link function, while not being limited to GPs with squared exponential kernels and rectangular domains.
\end{abstract}


\section{Introduction}
Estimating the intensity rate of discrete events over a continuous space is a common problem for real world applications such as modeling seismic activity \citep{ogata1998space}, neural data \citep{brillinger1988maximum}, forestry \citep{stoyan2000recent} and so forth. A particularly common approach is a Bayesian model based on a so--called Cox process \citep{cox1955some}. The observed events are assumed to be generated from a Poisson process, whose intensity function is modeled as another random process with a given prior probability measure. The problem of inference for such type of models has also attracted interest in the Bayesian machine learning community in recent years. \citet{moller1998log,brix2001spatiotemporal,cunningham2008inferring} assumed that the intensity function is sampled from a Gaussian Process (GP) prior~\citep{rasmussen2006gaussian}. However, to restrict the intensity function of the Poisson process to nonnegative values, a common strategy is to choose a nonlinear link function which takes the GP as its argument and returns a valid intensity. Based on the success of variational approximations to deal with complex Gaussian process models, the
inference problem for such Poisson models has attracted considerable interest in the machine learning community. 

While powerful black--box variational Gaussian inference algorithms are available which can be applied to arbitrary link--functions, the choice of link--functions is not only crucial for defining the prior over intensities but can also be important for the efficiency of variational inference. The 'standard' choice of 
Cox processes with an exponential link function was treated in \citep{hensman2015mcmc}.
However, variational Gaussian inference for this link function has the disadvantage 
that the posterior variance becomes decoupled from the observations~\citep{lloyd2015variational}.\footnote{\citet{samo2015scalable} propose an efficient approximate sampling scheme.} An interesting choice is the quadratic link function of \citep{lloyd2015variational} for which integrations over the data domain, which are necessary for sparse GP inference, can be (for specific kernel) computed analytically.\footnote{For a frequentist nonparametric approach to this model, see \citep{flaxman2017poisson}. For a Bayesian extension see~\citep{walder17fast}.} For both models, the minimisation of the variational free energies is performed by gradient descent techniques.

In this paper we will deal with approximate inference for a model with a sigmoid link--function.  This model was introduced by \citep{adams2009tractable} together with a MCMC sampling algorithm which was further improved by \citep{gunter2014efficient} and~\citep{teh2011gaussian}.~\citet{kirichenko2015optimality} have shown that the model has favourable (frequentist) theoretical properties provided priors and hyperparameters are chosen appropriately. In contrast to a direct variational Gaussian approximation for the posterior distribution of the latent function, we will introduce an alternative type of variational approximation which is specially designed for the {\em sigmoidal Gaussian Cox process}. We build on recent work on Bayesian logistic regression by data augmentation with P\'olya--Gamma random variables~\citep{polson2013bayesian}. This approach was already used in combination with GPs \citep{linderman2015dependent, wenzel2017scalable}, for stochastic processes in discrete time~\citep{linderman2017bayesian}, and for jump processes~\citep{donner2017inverse}. We extend this method to an augmentation by a latent, marked Poisson process, where the marks are distributed according to a P\'olya--Gamma distribution.\footnote{For a different application of marked Poisson processes, see~\citep{lloyd2016latent}.} In this way, the augmented likelihood becomes conjugate to a GP distribution. Using a combination of a mean--field variational approximation together with sparse GP approximations \citep{csato2002sparse,csato2002phd,titsias2009variational} we obtain explicit analytical variational updates leading to fast inference. In addition, we show that the same augmentation can be used for the computation of the maximum a posteriori (MAP) estimate by an expectation--maximisation (EM) algorithm. With this we obtain a Laplace approximation to the non--augmented posterior.

The paper is organised as follows: In section~\ref{sec:inference problem}, we introduce the sigmoidal Gaussian Cox process model and its transformation by the variable augmentation. In section~\ref{sec:inference}, we derive a variational mean field method and an EM--algorithm to obtain the MAP estimate, followed by the Laplace approximation of the posterior. Both methods are based on a sparse GP approximation to make the infinite dimensional problem tractable. In section~\ref{sec:results}, we demonstrate the  performance of our method on synthetic datasets and compare with the results of a Monte Carlo sampling method for the model and the variational approximation of~\citet{hensman2015mcmc}, which we modify to solve the Cox--process model with the scaled sigmoid link function. Then we compare our method to the state-of-the-art inference algorithm~\citep{lloyd2015variational} on artificial and real datasets with up to $10^4$ observations. Section~\ref{sec:discussion} presents a discussion and an outlook.

\section{The Inference problem}\label{sec:inference problem}

We assume that $N$ events $\dataset = \set{\bs{x}_n}_{n=1}^N$ are generated by a Poisson process. Each point $\bs{x}_n$ is a $d$--dimensional vector in the compact domain $\X\subset \mathbb{R}^d$. The goal is to infer the varying {\it intensity function} $\Lambda(\bs{x})$ (the mean measure of the process) for all $\bs{x}\in\X$ based on the likelihood 
\begin{equation*}\label{eq:poisson_likelihood}
L(\dataset\vert \Lambda) = \exp\left(-\int_{\X}\Lambda(\bx)d\bx\right)\prod_{n=1}^N\Lambda(\bx_n),
\end{equation*}
which is equal (up to a constant)  to the density of a  Poisson process having intensity
$\Lambda$ (see Appendix~\ref{app:variational inference} and \citep{konstantopoulos2011radon}) with respect to a Poisson process with unit intensity. In a Bayesian framework, a prior over the intensity makes $\Lambda$
a random process. Such a doubly stochastic point process is called {\it Cox process} \citep{cox1955some}. 
Since one needs $\Lambda(\bx)\geq 0$,  \citet{adams2009tractable} suggested a reparametrization of the intensity function by $\Lambda(\bx)=\lambda\sigma(g(\bx))$, where $\sigma(x)=(1+e^{-x})^{-1}$ is the sigmoid function and $\lambda$ is the maximum intensity rate. Hence, the intensity $\Lambda(\bx)$ is positive everywhere, for any arbitrary function $g(\bx): \X \rightarrow \mathbb{R}$ and the inference problem is to determine this function. Throughout this work we assume that $g(\cdot)$ will be modelled as a GP \citep{rasmussen2006gaussian} and the resulting process is  called {\it sigmoidal Gaussian Cox process}.
 The likelihood for $g$ becomes
\begin{equation}\label{eq:sigmoid poisson likelihood}
L(\dataset \vert g,\lambda) = \exp\left(-\int_{\X}\lambda\sigma(g(\bx))d\bx\right)\prod_{n=1}^N\lambda\sigma(g_n),
\end{equation}
where $g_n\doteq g(\bx_n)$. For Bayesian inference we define a GP prior measure $P_{\rm GP}$ with zero mean and covariance kernel $k(\bx,\bx^\prime):\X\times\X \rightarrow \R^+$. $\lambda$ has as prior density (with respect to the ordinary Lebesgue measure) $p(\lambda)$ which we take to be a Gamma density with shape-, and rate parameter $\alpha_0$ and $\beta_0$, respectively. Hence, for the prior we get the product measure $dP_{\rm prior}= dP_{\rm GP} \times p(\lambda)d\lambda$. The posterior density $\bs{p}$ (with respect to the prior measure) is given by
\begin{equation}\label{eq:posterior density}
\bs{p}(g,\lambda\vert \dataset) \doteq \frac{dP_{\rm posterior}}{dP_{\rm prior}}(g,\lambda\vert \dataset) = \frac{L(\dataset \vert g,\lambda)}{\EE{P_{\rm prior}}{L(\dataset \vert g,\lambda)}}.
\end{equation}
The normalising expectation in the denominator on the right hand side is with respect to the probability measure $P_{\rm prior}$. To deal with the infinite dimensionality of GPs and Poisson processes we require a minimum of extra notation. We introduce densities or {\em Radon--Nikod\'ym derivatives } such as defined in Equation~\eqref{eq:posterior density} (see Appendix~\ref{app:variational inference} or \citet{matthews2016sparse}) with respect to infinite dimensional measures by boldface symbols $\bs{p}(\bs{z})$. On the other hand, non--bold densities $p(\bs{z})$ denote densities in the `classical' sense, which means they are with respect to Lebesgue measure $d\bs{z}$. 

Bayesian inference for this model is known to be doubly intractable~\citep{murray2006mcmc}. The likelihood in Equation~\eqref{eq:sigmoid poisson likelihood} contains the integral of $g$ over the space $\X$ in the exponent and the normalisation of the posterior in Equation~\eqref{eq:posterior density} requires calculating expectation of Equation~\eqref{eq:sigmoid poisson likelihood}. In addition inference is hampered by the fact, that likelihood~\eqref{eq:sigmoid poisson likelihood} depends non--linearly on $g$ (through sigmoid and exponent of sigmoid). In the following we tackle this by an augmentation scheme for the likelihood, such that it becomes conjugate to a GP prior and we subsequently can derive an analytic form of a variational posterior given one simple mean field assumption (Section~\ref{sec:inference}).

\subsection{Data augmentation I: Latent Poisson process}
We will briefly introduce a data augmentation scheme by a latent Poisson process which forms the basis of 
the sampling algorithm of \citet{adams2009tractable}. We will then extend this method further to an 
augmentation by a {\em marked} Poisson process.
We focus on the exponential term in Equation~\eqref{eq:sigmoid poisson likelihood}. Utilizing the well known property of the sigmoid 
that $\sigma(x) = 1 - \sigma(-x)$ we can write
\begin{equation}\label{eq:exponent integral}
\begin{split}
\exp\left(-\int_{\X}\lambda\sigma(g(\bx))d\bx\right) = & \exp\left(-\int_{\X}
\left(1 - \sigma(-g(\bx))\right) \lambda d\bx \right).
\end{split}
\end{equation}
The left hand side has the form of a characteristic functional of a Poisson process. Generally, for a random set of points $\Pi_\calZ=\set{\bz_m;\bz_m\in \calZ}$ on a space $\calZ$ and with a function $h(\bs{z})$, this is defined as
\begin{equation}\label{eq:poisson characteristic functional}
\EE{P_\Lambda}{\prod_{\bs{z}_m\in\Pi_{\mathcal{Z}}} e^{h(\bs{z}_m)}} = \exp\left(-\int_{\mathcal{Z}}\left(1 - e^{h(\bs{z})}\right)\Lambda(\bs{z})d\bs{z}\right),
\end{equation}
where $P_\Lambda$ is the probability measure of a Poisson process with intensity $\Lambda(\bs{z})$. Equation~\eqref{eq:poisson characteristic functional} can be derived by Campbell's theorem (see Appendix~\ref{app:poisson process} and \citep[chap. 3]{kingman1993poisson}) and identifies a Poisson process uniquely. 

Setting $h(\bs{z}) = \ln \sigma(-g(\bs{z}))$, and $\mathcal{Z}=\X$, and combining Equation~\eqref{eq:exponent integral} and~\eqref{eq:poisson characteristic functional} we obtain the likelihood used by \citet[Eq. 4]{adams2009tractable}. However, in this work we make use of another augmentation, before invoking Campbell's theorem. This will result in a likelihood which is conjugate to the model priors and further simplifies inference.

\subsection{Data augmentation II: P\'olya--Gamma variables and marked Poisson process}
Following \citet{polson2013bayesian} we represent the inverse of the hyperbolic cosine as a scaled Gaussian mixture model 
\begin{equation}\label{eq:polya-gamma}
\cosh^{-b}(z/2)=\int_0^\infty e^{-\frac{z^2}{2}\omega}\PG(\omega\vert b,0) d\omega,
\end{equation}
where $\PG$ is a {\it P\'olya--Gamma} density (Appendix~\ref{app:polya gamma}). We further define the {\it tilted} P\'olya--Gamma density by
\begin{equation}\label{eq:tilted PG}
\PG(\omega\vert b,c)\propto e^{-\frac{c^2}{2}\omega}\PG(\omega\vert b,0),
\end{equation}
where $b>0$ and $c$ are parameters.  We will not need an explicit form of this density, since the subsequently derived inference algorithms will only require the first moments. Those can be obtained directly from the moment generating function, which can be calculated straightforwardly from Equation~\eqref{eq:polya-gamma} and~\eqref{eq:tilted PG} (see Appendix~\ref{app:polya gamma}). Equation~\eqref{eq:polya-gamma} allows us to rewrite the sigmoid function as
\begin{equation}\label{eq:PG sigmoid}
\sigma(z) = \frac{e^\frac{z}{2}}{2\cosh(\frac{z}{2})}= \int_{0}^\infty e^{f(\omega,z)}\PG(\omega\vert 1,0) d\omega,
\end{equation}
where we define 
\begin{equation}
f(\omega,z) \doteq \frac{z}{2} - \frac{z^2}{2}\omega - \ln 2. 
\end{equation}
Setting $z=-g(\bx)$ in Equation~\eqref{eq:exponent integral} and substituting Equation~\eqref{eq:PG sigmoid} we get
\begin{equation}\label{eq:marked poisson}
\begin{split}
& \exp\left(-\int_{\X} \lambda \left(1 - \sigma(-g(\bx))\right) d\bx \right)  = \exp\left(-\int_{\X\times \R^+}\left(1 - e^{f(\omega,-g(\bx))}\right) \;\PG(\omega\vert 1,0)\;  \lambda d\omega d\bx\right).
\end{split}
\end{equation}
Finally, we apply Campbell's theorem (Equation~\eqref{eq:poisson characteristic functional}) to Equation~\eqref{eq:marked poisson}. The space is a product space $\mathcal{Z}=\hat{\X}\doteq\X \times \R^+$ and the intensity $\Lambda(\bx,\omega) = \lambda \PG(\omega\vert 1, 0)$. This results in the final representation of the exponential in Equation~\eqref{eq:marked poisson}
\begin{equation}
\exp\left(-\int_{\hat{\X}}\left(1 - e^{f(\omega,-g(\bx))}\right) \Lambda(\bx,\omega)\;d\omega d\bx \right)
 = 
\EE{P_\Lambda}{\prod_{(\bx,\omega)_m\in \Pi_{\hat{\X}}} e^{f(\omega_m,-g_m)}}.
\end{equation}
Interestingly, the new Poisson process $\Pi_{\hat{\X}}$ with measure $P_\Lambda$ has the form of a {\it marked} Poisson process~\citep[chap. 5]{kingman1993poisson}, where the latent P\'olya-Gamma variables $\omega_m$ denote the `marks' being independent random variables at each location $\bx_m$. It is straightforward to sample such processes by first sampling the inhomogeneous Poisson process on domain $\X$ (for example by `thinning' a process with constant rate \citep{lewis1979simulation,adams2009tractable}) and then drawing a mark $\omega$ on each event independently from the density $\PG(\omega\vert 1,0)$. 

Finally, using the P\'olya--Gamma augmentation also for the discrete likelihood factors corresponding to the observed events in Equation~\eqref{eq:sigmoid poisson likelihood} we obtain the following joint likelihood of the model
\begin{equation}\label{eq:augmented likelihood}
\begin{split}
L(\dataset, \bomega_N, \Pi_{\hat{\X}}\vert g, \lambda) \doteq &
\frac{dP_{\rm joint}}{dP_{\rm aug}} (\dataset, \bomega_N, \Pi_{\hat{\X}}\vert g, \lambda) \\
= & \;  \prod_{(\bx,\omega)_m\in\Pi_{\hat{\X}}} e^{f(\omega_m,-g_m)}  \prod_{n=1}^N \lambda e^{f(\omega_n,g_n)},
\end{split}
\end{equation}
where we define the prior measure of augmented variables as $P_{\rm aug} = P_{\Lambda} \times P_{\bomega_N}$ and where $\bomega_N=\set{\omega_n}_{n=1}^N$ are the P\'olya--Gamma variables for the observations $\dataset$ with the prior measure $dP_{\bomega_N} = \prod_{n=1}^Np(\omega_n\vert 1, 0)d\omega_n$. This augmented representation of the likelihood contains the function $g(\cdot)$ only linearly and quadratically in the exponents and is thus conjugate to the GP prior of $g(\cdot)$. Note that the original likelihood in Equation~\eqref{eq:sigmoid poisson likelihood} can be recovered by $\EE{P_{\rm aug}}{L(\dataset, \bomega_N, \Pi_{\hat{\X}} \vert g, \lambda)} = L(\dataset \vert g, \lambda)$.

\section{Inference in the augmented space}\label{sec:inference}
Based on the augmentation we
define a posterior density for the joint model with respect to the product measure $P_{\rm prior} \times P_{\rm aug}$
\begin{equation}\label{eq:posterior}
\begin{split}
\bs{p}(\bomega_N, \Pi_{\hat{\X}}, g, \lambda \vert \dataset) \doteq & \;
\frac{dP_{\rm posterior}}{d(P_{\rm prior} \times P_{\rm aug})} (\bomega_N, \Pi_{\hat{\X}}, g, \lambda \vert \dataset)\\
= &\; \frac{L(\dataset, \bomega_N, \Pi_{\hat{\X}} \vert g, \lambda)}{L(\dataset)},
\end{split}
\end{equation}
where the denominator is the marginal likelihood $L(\dataset) = \EE{P_{\rm prior} \times P_{\rm aug}}{L(\dataset, \bomega_N, \Pi_{\hat{\X}}\vert g, \lambda)}$.
The posterior density of Equation~\eqref{eq:posterior} could be sampled using Gibbs sampling with explicit, tractable conditional densities. Similar to the variational approximation in the next section, one can show that the conditional measure of the point sets $\Pi_{\hat{\X}}$ and the variables $\bomega_N$, given the 
function $g(\cdot)$ and maximal intensity $\lambda$ is a product of a specific marked Poisson process and independent (tilted) P\'olya--Gamma densities. On the other hand, the distribution over function $g(\cdot)$ conditioned on $\Pi_{\hat{\X}}$ and $\bomega_N$ is a Gaussian process. Note, however, one needs to sample this GP only at the finite points $\bx_m$ in the random set $\Pi_{\hat{\X}}$ and the fixed set $\dataset$.

\subsection{Variational mean--field approximation}\label{sec:variational}
For variational inference one assumes that the desired posterior probability measure belongs to a family of measures for which the inference problem is tractable. Here we make a simple structured mean field assumption in order to fully utilise its conjugate structure: We approximate the posterior measure by
\begin{equation}\label{eq:variational posterior measure}
P_{\rm posterior}(\bomega_N, \Pi_{\hat{\X}}, g, \lambda\vert \dataset)   \approx Q_1(\bomega_N, \Pi_{\hat{\X}}) \times Q_2(g,\lambda),
\end{equation}
meaning that the dependencies between the P\'olya--Gamma variables $\bomega_N$ and the marked Poisson process $\Pi_{\hat{\X}}$ on the one hand, and the function $g$ and the maximal intensity $\lambda$
on the other hand, are neglected. As we will see in the following, this simple mean--field assumption allows us to derive the posterior approximation analytically. 

The variational approximation is optimised by minimising the Kullback--Leibler divergence between exact and approximated posteriors. This is equivalent to maximising the  lower bound on the marginal likelihood of the observations
\begin{equation}\label{eq:lower bound}
\mathcal{L}(\bs{q}) = \EE{Q}{\log \left\{\frac{L(\dataset, \bomega_N, \Pi_{\hat{\X}} \vert g, \lambda)}
{\bs{q}_1(\bomega_N, \Pi_{\hat{\X}})\bs{q}_2(g,\lambda)}
\right\}} \leq \log L(\dataset),
\end{equation}
where $Q$ is the probability measure of the variational posterior in Equation~\eqref{eq:variational posterior measure} and we introduced approximate likelihoods
\begin{equation}
\bs{q}_1(\bomega_N, \Pi_{\hat{\X}})\doteq \frac{dQ_1}{dP_{\rm aug}}
(\bomega_N, \Pi_{\hat{\X}}), \qquad
\bs{q}_2(g,\lambda) \doteq \frac{dQ_2}{dP_{\rm prior}}(g,\lambda).
\end{equation}

Using standard arguments for mean field variational inference \citep[chap. 10]{bishop2006pattern} and Equation~\eqref{eq:variational posterior measure}, one can then show
that the optimal factors satisfy
\begin{equation}\label{eq:optimal first factor}
\ln \bs{q}_1\left(\bomega_N, \Pi_{\hat{\X}}\right) =  \EE{Q_2}{\log L(\dataset, \bomega_N, \Pi_{\hat{\X}} \vert
g, \lambda)  } + \mbox{const.}
\end{equation}
and
\begin{equation}\label{eq:optimalq2}
\ln \bs{q}_2(g,\lambda) =  \EE{Q_1}{\log L(\dataset, \bomega_N, \Pi_{\hat{\X}} \vert g, \lambda) } + \mbox{const.}\;,
\end{equation}
respectively. These results lead to an iterative scheme for optimising $\bs{q}_1$ and $\bs{q}_2$ in order to increase
the lower bound in Equation~\eqref{eq:lower bound} in every step. 
From the structure of the likelihood one derives two further factorisations:
\begin{align}\label{eq:factorizing form1}
& \bs{q}_1(\bomega_N, \Pi_{\hat{\X}}) = \bs{q}_1(\bomega_N) \bs{q}_1(\Pi_{\hat{\X}}), \\
\label{eq:factorizing form2}
& \bs{q}_2(g,\lambda) = \bs{q}_2(g) \bs{q}_2(\lambda),
\end{align}
where the densities are defined with respect to the measures $dP(\bomega_N),\; dP_\Lambda,\; dP_{\rm GP}$, and $p(\lambda)d\lambda$, respectively. The subsequent section describes these updates explicitly.
\paragraph{Optimal P\'olya--Gamma density}
Following Equation~\eqref{eq:optimal first factor} and \eqref{eq:factorizing form1} we obtain
\begin{equation}
\begin{split}
\bs{q}_1(\bomega_N) = \prod_{n=1}^N \frac{\exp\left(-\frac{c_1^{(n)}}{2}\omega_n\right)}{\cosh^{-1}\left(c_1^{(n)}/2\right)} = \prod_{n=1}^N \frac{\PG\left(\omega_n \vert 1, c_{1}^{(n)}\right)}{\PG\left(\omega_n \vert 1, 0\right)},
\end{split}
\end{equation}
where the factors are {\it tilts} of the prior P\'olya-Gamma densities (see Equation~\eqref{eq:tilted PG} and Appendix~\ref{app:polya gamma}) with $c_1^{(n)}=\sqrt{\EE{Q_2}{g_n^2}}$. By simple density transformation we obtain the density with respect to the Lebesgue measure as
\begin{equation}\label{eq:polya gamma obs}
q_1(\bomega_N) = \bs{q}_1(\bomega_N)\left\vert \frac{dP_{\bomega_N}}{d\bomega_N} \right\vert =  \prod_{n=1}^N \PG\left(\omega_n \vert 1, c_{1}^{(n)}\right),
\end{equation}
being a product of {\it tilted} P\'olya--Gamma densities.
\paragraph{Optimal Poisson process}
Using Equation~\eqref{eq:optimal first factor}  and~\eqref{eq:factorizing form1} we obtain
\begin{equation}\label{eq:optimal marked process}
\begin{split}
\bs{q}_1(\Pi_{\hat{\X}})  = & \frac{\prod_{(\bx,\omega)_m\in\Pi_{\hat{\X}}} e^{\EE{Q_2}{f(\omega_m,-g_m)}}\lambda_{\scriptstyle 1}}{\exp\left(\int_{\hat{\X}} \left(e^{\EE{Q_2}{f(\omega,-g(\bx))}}-1\right)
\lambda_{1}\PG(\omega\vert 1,0) d\bx d\omega\right)},
\end{split}
\end{equation}
with $\lambda_{1} \doteq e^{\EE{Q_2}{\log \lambda^*}}$. Note, that $\EE{Q_2}{f(\omega_m,-g_m)}$ involves the expectations $\EE{Q_2}{g_m}$ and $\EE{Q_2}{(g_m)^2}$. One can show, that Equation~\eqref{eq:optimal marked process} is again a marked Poisson process with intensity
\begin{equation}\label{eq:posterior mean measure}
\begin{split}
\Lambda_{1}(\bx,\omega) = & \lambda_{1}\frac{\exp\left(-\frac{\EE{Q_2}{g(\bx)}}{2} \right)}{2\cosh\left(\frac{c_{1}(\bx)}{2}\right)} 
\PG\left(\omega \vert 1, c_{1}(\bx)\right) \\
= & \lambda_1 \sigma(-c_1(\bx))\exp\left(\frac{c_1(\bx) - \EE{Q_2}{g(\bx)}}{2}\right)\PG\left(\omega \vert 1, c_{1}(\bx)\right)
\end{split}
\end{equation}
where $c_{1}(\bx) = \sqrt{\EE{Q_2}{g(\bx)^2}}$ (for a proof see Appendix~\ref{app:optimal posterior poisson}).


\paragraph{Optimal Gaussian process}
From Equation~\eqref{eq:optimalq2} and~\eqref{eq:factorizing form2}
we obtain the optimal approximation of the posterior likelihood (note that this is defined relative to GP prior) \begin{equation}
\bs{q}_2(g) \propto e^{U(g)},
\end{equation}
where the effective log--likelihood is given by
\begin{equation}
U(g) =  \EE{Q_1}{\sum_{(\bx,\omega)_m\in\Pi_{\hat{\X}}}f(\omega_m,-g_m)}
+  \sum_{n=1}^N \EE{Q_1}{f(\omega_n,g(\bx_n))}.
\end{equation}
The first expectation is over the variational Poisson process $\Pi_{\hat{\X}}$ and the second one over the P\'olya--Gamma variables $\bomega_N$. These can be easily evaluated (see Appendix~\ref{app:poisson process}) and one finds
\begin{equation}\label{eq:U function}
U(g) =  -\frac{1}{2}\int_\X A(\bx) g(\bx)^2 d\bx + \int_\X B(\bx)g(\bx)d\bx,
\end{equation}
with
\begin{align}
A(\bx) = &  \sum_{n=1}^N \EE{Q_1}{\omega_n}\delta(\bx - \bx_n) + \int_0^\infty \omega \Lambda_{1}(\bx,\omega) d\omega,\\
B(\bx) = & \frac{1}{2} \sum_{n=1}^N \delta(\bx - \bx_n) - \frac{1}{2}\int_0^\infty \Lambda_1(\bx,\omega)
d\omega ,
\end{align}
where $\delta(\cdot)$ is the Dirac delta function. The expectations and integrals over $\omega$ are 
\begin{align}
&\EE{Q_1}{\omega_n} = \frac{1}{2c_1^{(n)}}\tanh\left(\frac{c_1^{(n)}}{2}\right),\\
&\int_0^\infty \Lambda_{1}(\bx,\omega) d\omega = \lambda_1 \sigma(-c_1(\bx))\exp\left(\frac{c_1(\bx) - \EE{Q_2}{g(\bx)}}{2}\right) \doteq \Lambda_1(\bx),  \\
&\int_0^\infty \omega \Lambda_{1}(\bx,\omega) d\omega = \frac{1}{2 c_{1}(\bx)}\tanh\left(\frac{c_{1}(\bx)}{2}\right)\Lambda_1(\bx).
\end{align}
The resulting variational distribution defines a Gaussian process. Because of the mean--field assumption the integrals in Equation~\eqref{eq:U function} do not require integration over random variables, but only solving two deterministic integrals over space $\X$. However, those integrals depend on function $g$ over the entire space and it is not possible for a general kernel to compute the marginal posterior density at an input $\bx$ in closed form. 
For specific GP kernel operators, which are the inverses of differential operators, a solution in terms of 
linear partial differential equations would be possible. This could be of special interest for 
one--dimensional problems where Matern kernels with integer parameters~\citep{rasmussen2006gaussian}
fulfill this condition. Here, the problem becomes equivalent to inference for  a (continuous time) Gaussian hidden Markov model and could be solved by performing a forward--backward algorithm~\citep{solin2016stochastic}. This would reduce the computations to the solution of ordinary differential equations. We will discuss details of such an approach elsewhere. To deal with general kernels we will resort instead to a 
the well known variational sparse GP approximation with inducing points.
\paragraph{Optimal sparse Gaussian process}
The sparse variational Gaussian approximation follows the standard approach~\citep{csato2002sparse,csato2002phd,titsias2009variational} and its generalisation to a continuum likelihood \citep{batz2018approximate,matthews2016sparse}. For completeness, we repeat the
derivation here and more detailed in Appendix E. We approximate $\bs{q}_2(g)$ by a sparse 
likelihood GP $\bs{q}_2^s(g)$ with respect to the GP prior
\begin{equation}\label{eq:sparse GP model}
\frac{dQ_2^s}{dP}(g) = \bs{q}_2^s(\bg_s),
\end{equation}
which depends only on a finite dimensional vector
of function values $\bs{g}_s = (g(\bx_1),\ldots,g(\bx_L))^\top$ at a set of {\it inducing points} $\set{\bx_l}_{l=1}^L$. With this approach it is again possible to marginalise out exactly all the infinitely many function values outside of the set of inducing points. 
The sparse likelihood $\bs{q}_2^s$ is optimised by minimising the Kullback--Leibler divergence
\begin{equation}
{\rm D}_{\rm KL}(Q_2^s\Vert Q_2) = \EE{Q_2^s}{\log\frac{\bs{q}_2^s(g)}{\bs{q}_2(g)}}.
\end{equation}
A short computation (Appendix~\ref{app:sparse GP}) shows that 
\begin{equation}
q_2^s(\bg_s) \propto e^{U^s(\bg_s)}\qquad {\rm with}\ U^s(\bs{g}_s) = \EE{P(g\vert\bg_s)}{U(g)}, 
\end{equation}
where the conditional expectation is with respect to the GP prior measure given the function $\bg_s$ at the inducing points. The explicit calculation requires the conditional expectations of $g(\bx)$ and of $(g(\bx))^2$. We get
\begin{equation}\label{eq:expected g}
\EE{P(g\vert\bg_s)}{g(\bx)} = \boldsymbol{k}_s(\bx)^\top\; K_s^{-1}\boldsymbol{g}_s,
\end{equation}
where $\bks(\bx)=(k(\bx,\bx_1),\ldots,k(\bx,\bx_L))^\top$ and $K_s$ is the kernel matrix between inducing points.
For the second expectation, we get 
\begin{equation}\label{eq:expected squared g}
\EE{P(g\vert\bg_s)}{g^2(\bx)} = \left(\EE{P(g\vert\bg_s)}{g(\bx)}\right)^2 + \mbox{const.}
\end{equation}
The constant equals the conditional variance of $g(\bx)$ which does not depend on the sparse set $\bs{g}_s$, 
but only on the locations of the sparse points. Because we are dealing now with a finite problem we can define the `ordinary' posterior density of the GP at the inducing points
with respect to the Lebesgue measure $d\bg_s$. From Equation~\eqref{eq:U function},~\eqref{eq:expected g}, and~\eqref{eq:expected squared g}, we conclude that the sparse posterior at the inducing variables is a multivariate Gaussian density
\begin{equation}\label{eq:GP posterior}
q_2^{s}(\bs{g}_s) = \mathcal{N}(\bs{\mu}_2^s,\Sigma_2^s),
\end{equation}
with the covariance matrix given by 
\begin{equation}\label{eq:posterior GP cov}
\Sigma_2^s = \left[K_s^{-1}\int_{\X} A(\bx)\bs{k}_s(\bx)\bs{k}_s(\bx)^\top d\bx\; K_s^{-1} + K_s^{-1}\right]^{-1},
\end{equation}
and the mean
\begin{equation}\label{eq:posterior GP mean}
\boldsymbol{\mu}_2^s = \Sigma_2^s\left(K_s^{-1}\int_{\X} B(\bx)\bs{k}_s(\bx)d\bx\right).
\end{equation}
In contrast to other variational approximations (see for example~\citep{lloyd2015variational,hensman2015mcmc})
we obtain a closed analytic form of the variational posterior mean and covariance which holds for
arbitrary GP kernels. However, these results depend on finite dimensional integrals over the space $\X$
which cannot be computed analytically. This is different to the sparse approximation
for the Poisson model with square link function~\citep{lloyd2015variational}, where similar integrals in the case of the squared exponential kernel can be obtained analytically. Hence, 
we resort to a simple Monte--Carlo integration, where {\it integration points} are sampled uniformly on $\X$ as
\begin{equation}
I_F = \int_\X F(\bx) d\bx \approx \frac{|\X|}{R}\sum_{r=1}^R F(\bx_r).
\end{equation}
The set of integration points $\set{\bx_r}_{r=1}^R$ is drawn uniformly from the space $\X$.

Finally, from Equation~\eqref{eq:sparse GP model} and~\eqref{eq:GP posterior} we obtain the mean function and the variance of the sparse approximation for every point $\bx\in \X$, which is 
\begin{equation}\label{eq:pred GP mean}
\mu_{2}(\bx)=\EE{Q_2}{g(\bx)} = \bs{k}_s(\bx)^\top K_s^{-1}\boldsymbol{\mu}_2^s,
\end{equation}
and variance
\begin{equation}
\begin{split}\label{eq:pred GP var}
\left(s_{2}(\bx)\right)^2 = k(\bx,\bx) - \bs{k}_s(\bx)^\top K_s^{-1}\left(\mathbf{I} - \Sigma_2^s K_s^{-1}\right)\bs{k}_s(\bx),
\end{split}
\end{equation}
where $\mathbf{I}$ is the identity matrix.

\paragraph{Optimal density for maximal intensity $\lambda$}
From Equation~\eqref{eq:optimalq2} we identify the optimal density as a Gamma density
\begin{equation}\label{eq:optimal lambda density}
q_2(\lambda) = \rm{Gamma}(\lambda\vert \alpha_{2}, \beta_{2}) = \frac{\beta_{2}^{\alpha_{2}}(\lambda)^{\alpha_{2}-1}e^{-\beta_{2}\lambda}}{\Gamma(\alpha_{2})},
\end{equation}
where $\alpha_{2}= N + \EE{Q_1}{\boldsymbol{1}_{\Pi}(\bx)} + \alpha_0$, $\beta_{2}=\beta_0 + \int_\X d\bx$ and $\Gamma(\cdot)$ is the gamma function. $\boldsymbol{1}_{\Pi}(\bx)$ denotes the indicator function being $1$ if $\bx\in \Pi$ and $0$ otherwise and the integral is again solved by Monte Carlo integration. This defines the required expectations for updating $q_1$ by $\EE{Q_2}{\lambda} = \frac{\alpha_{2}}{\beta_{2}}$ and 
$\EE{Q_2}{\log\lambda} = \psi(\alpha_{2}) - \log \beta_{2}$,
where $\psi(\cdot)$ is the digamma function.

\paragraph{Hyperparameters} Hyperparameters of the model are (i)  the covariance parameters $\btheta$ of the GP, (ii) the locations of the inducing points $\set{\bx_l}_{l=1}^L$, and (iii) the prior parameters $\alpha_0,\beta_0$ for the maximal intensity $\lambda$. The covariance parameters (i) $\btheta$ are optimised by gradient ascent following the gradient of the lower bound in Equation~\eqref{eq:lower bound} with respect to $\btheta$ (Appendix~\ref{app:lower bound}). As gradient ascent algorithm we employ the ADAM algorithm~\citep{kingma2014adam}. We perform always one step after the variational posterior $q$ is updated as described before. (ii) The locations of the sparse GP $\set{\bx_l}_{l=1}^L$ could in principle be optimised as well, but we keep them fixed and position them on a regular grid over the space $\X$. From this choice it follows that $K_s$ is a Toeplitz matrix, when the kernel is translationally invariant. This could be inverted in $\mathcal{O}(L (\log L)^2)$ instead of  $\mathcal{O}(L^3)$ operations~\citep{press1989numerical} but we do not employ this fact. Finally, (iii) the value for prior parameters $\alpha_0$ and $\beta_0$ are chosen such that $p(\lambda)$ has a mean twice and standard deviation once the intensity one would expect for a homogeneous Poisson Process observing $\dataset$. The complete variational procedure is outlined in Algorithm~\ref{alg:VB}.

\begin{algorithm}\label{alg:VB}
    \SetKwInOut{Init}{Init}
    \SetKw{Updatehyper}{Update kernel parameters with gradient ascent}
    \SetKwBlock{Updateqone}{Update $q_1$}{}
    \SetKwBlock{Updateqtwo}{Update $q_2$}{}

    \Init{$\EE{Q}{g(\bx)},\EE{Q}{(g(\bx))^2}$ at $\dataset$ and integration points, and $\EE{Q}{\lambda},\EE{Q}{\log \lambda}$}
    \While{$\mathcal{L}$ not converged}
      {
        \Updateqone{
        {\bf PG distributions at observations}: $q_1(\bomega_N)$ with Eq.~\eqref{eq:polya gamma obs}\\
        {\bf Rate of latent process}: $\Lambda_1(\bx,\omega)$ at integration points with Eq.~\eqref{eq:posterior mean measure}        
      		       }
        \Updateqtwo{
      {\bf Sparse GP distribution}: $\Sigma_2^s,\bmu_2^s$ with Eq.~\eqref{eq:posterior GP cov},~\eqref{eq:posterior GP mean}\\
      {\bf GP at $\dataset$ and integration points}: $\EE{Q_2}{g(\bx)},\EE{Q_2}{(g(\bx))^2}$ with Eq.~\eqref{eq:pred GP mean},~\eqref{eq:pred GP var}\\
      {\bf Gamma-distribution of $\lambda$}: $\alpha_2,\beta_2$ with Eq.~\eqref{eq:optimal lambda density}
      }
      \Updatehyper
      }
      
    \caption{Variational Bayes algorithm for sigmoidal Gaussian Cox process.}
\end{algorithm}

\subsection{Laplace approximation}\label{subsec:laplace}
In this section we will show that our variable augmentation method is well suited for computing a
Laplace approximation~\citep[chap. 4]{bishop2006pattern} to the joint posterior of the GP function $g(\cdot)$ and the 
maximal intensity $\lambda$ as an alternative to the previous variational scheme. To do so we need the maximum a posteriori (MAP) estimate (equal to the mode of the posterior distribution) and a second order Taylor expansion around this mode. The augmentation method will be used to compute the MAP estimator
iteratively using an EM algorithm.

\paragraph{Obtaining the MAP estimate} 
In general, a proper definition of the posterior mode would be necessary, because the 
GP posterior is over a space of functions, which is an infinite dimensional object and does not have a 
density with respect to Lebesgue measure. A possibility to avoid this problem would be to
discretise the spatial integral in the likelihood and to approximate the posterior by a multivariate 
Gaussian density
for which the mode can then be computed by setting the gradient equal to zero.
In this paper, we will use a different approach which defines the mode directly 
in function space and allows us to utilise the sparse
GP approximation developed previously for the computations.  A mathematically proper way
would be to derive the MAP estimator by maximising a properly penalised log--likelihood. As discussed e.g.
in \citet[chap. 6]{rasmussen2006gaussian} for GP models with likelihoods which depend on finitely many 
inputs only, this penalty is given by the squared reproducing kernel Hilbert space (RKHS) norm 
that corresponds to the GP kernel. Hence, we would have
\begin{equation}\label{eq:MAP problem} 
\begin{split}
(g^*,\lambda^*) = &  {\rm argmin}_{g\in \calH_k, \lambda} \left\lbrace - \ln L(\dataset\vert g, \lambda) -\ln p(\lambda)
+ \frac{1}{2}\Vert g \Vert_{\calH_k}^2 \right\rbrace,
\end{split}
\end{equation}
where $\Vert g \Vert_{\calH_k}^2$ is the RKHS norm for the kernel $k$. 
This penalty term can be understood as a proper generalisation of a Gaussian log--prior density to function space.
We will not give a formal definition
here  but work on a more heuristic level in the following. Rather than attempting 
a direct optimisation, we will use an EM algorithm instead, applying the variable augmentation with the Poisson process and P\'olya--Gamma variables introduced in the previous sections. 
In this case, the likelihood part of the resulting '$\Q$--function' 
\begin{equation}\label{eq:Qfunction}
\Q((g,\lambda)\vert (g,\lambda)^{\rm old}) \doteq \EE{P(\bomega_N,\Pi_{\hat{\X}}\vert (g,\lambda)^{\rm old})}{\ln L(\dataset,\bomega_N,\Pi_{\hat{\X}}\vert g,\lambda)} + \ln p(\lambda) - \frac{1}{2}\Vert g \Vert_{\calH_k}^2,
\end{equation}
that needs to be maximised in the M--step becomes (as in the variational approach before) the likelihood
of {\em a Gaussian model} in the GP function $g$. 
Hence, we can argue that the function $g$ which maximises $\Q$ is equal to the {\em posterior mean} of the resulting Gaussian model and can be computed
without discussing the explicit form of the RKHS norm.  

The conditional probability measure $P(\bomega_N,\Pi_{\hat{\X}}\vert (g,\lambda)^{\rm old})$ is easily obtained similar to the optimal measure $Q_1$ by not averaging over $g$ and $\lambda$. This gives us straightforwardly the density
\begin{equation}
\bs{p}(\bomega_N,\Pi_{\hat{\X}}\vert (g,\lambda)^{\rm old}) = \bs{p}(\bomega_N\vert (g,\lambda)^{\rm old})\bs{p}(\Pi_{\hat{\X}}\vert (g,\lambda)^{\rm old}).
\end{equation}
The first factor is
\begin{equation}
p(\bomega_N\vert  (g,\lambda)^{\rm old}) = \bs{p}(\bomega_N\vert  (g,\lambda)^{\rm old})\left\vert\frac{dP_{\bomega_N}}{d\bomega_N} \right\vert =\prod_{n=1}^N \PG\left(\omega_n \vert 1, \tilde{c}_n\right),
\end{equation}
with $\tilde{c}_n=|g_n^{\rm old}|$. The latent point process $\Pi_{\hat{\X}}$ is again a Poisson process density 
\begin{equation}
\bs{p}(\Pi_{\hat{\X}}\vert (g,\lambda)^{\rm old}) =\frac{dP_{\tilde{\Lambda}}}{dP_\Lambda}(\Pi_{\hat{\X}}\vert (g,\lambda)^{\rm old}),
\end{equation}
where the intensity is
\begin{equation}
\tilde{\Lambda}(\bx,\omega) = \lambda^{\rm old}\sigma(-g^{\rm old}(\bx))\PG\left(\omega \vert 1, \tilde{c}(\bx)\right),
\end{equation}
with $\tilde{c}(\bx) = |g^{\rm old}(\bx)|$. 
The first term in the $\Q$--function is 
\begin{equation}\label{eq:MAP log likelihood}
\begin{split}
U(g,\lambda) \doteq &\; \EE{P(\bomega_N,\Pi_{\hat{\X}}\vert (g,\lambda)^{\rm old})}{\ln L(\dataset,\bomega_N,\Pi_{\hat{\X}}\vert g,\lambda)} \\
= & - \frac{1}{2}\int_{\X} \tilde{A}(\bx)g(\bx)^2 d\bx + \int_\X \tilde{B}(\bx) g(\bx) d\bx,
\end{split}
\end{equation}
with
\begin{align}
\tilde{A}(\bx) & = \sum_{n=1}^N \EE{P(\omega_n\vert (g,\lambda)^{\rm old})}{\omega_n}\delta(\bx - \bx_n) + \int_{0}^\infty \EE{P(\omega\vert (g,\lambda)^{\rm old})}{\omega}\tilde{\Lambda}(\bx,\omega)d\omega, \\
\tilde{B}(\bx) & = \frac{1}{2}\sum_{n=1}^N \delta(\bx - \bx_n) - \frac{1}{2}\int_{0}^\infty \tilde{\Lambda}(\bx,\omega)d\omega.
\end{align}
We have already tackled almost identical log--likelihood expressions in Section~\ref{sec:variational} (see Equation~\eqref{eq:U function}). While for specific priors (with precision kernels given by differential operators) an exact treatment in terms of solutions of ODEs or PDEs is possible, we will again resort to the sparse GP approximation instead. The sparse version $U^s(\bg_s,\lambda)$ is obtained by replacing 
$g(\bx) \rightarrow \EE{P(g\vert \bg_s)}{g(\bx)}$ in $U(g,\lambda)$.
From this we obtain the sparse $\Q$--function as
\begin{equation}\label{eq:sparse Qfunction}
\Q^{s}((\bg_s,\lambda)\vert (\bg_s,\lambda)^{\rm old}) \doteq U^s(\bg_s,\lambda) + \ln p(\lambda) - \frac{1}{2}\bg_s^\top K^{-1}_s \bg_s.
\end{equation}
The function values $\bg_s$ and the maximal intensity $\lambda$ that maximise Equation~\eqref{eq:sparse Qfunction} can be found analytically by solving
\begin{equation}\label{eq:Mstep}
\frac{\partial \Q^s}{\partial \bg_s} = \bs{0}\;\rm{and}\;\frac{\partial \Q^s}{\partial \lambda} = 0. 
\end{equation}
The final MAP estimate is obtained after convergence of the EM algorithm and 
the desired sparse MAP solution for $g(x)$ is given by
(see Equation~\eqref{eq:pred GP mean})
\begin{equation}
g_{MAP}(\bx)= \bs{k}_s(\bx)^\top K_s^{-1}\bg^s
\end{equation}
As for the variational scheme, integrals over the space $\X$ are approximated by Monte--Carlo integration.
An alternative derivation of the sparse MAP solution can be based on 
 restricting the minimisation of \eqref{eq:Qfunction} to functions which are linear combinations of kernels 
 centred at the inducing points and using the definition of the RKHS norm
(see \citep[chap. 6]{rasmussen2006gaussian}). 

\paragraph{Sparse Laplace posterior} To complete the computation of the Laplace approximation, we need
to evaluate the quadratic fluctuations around the MAP solution. We will also do this with the previously obtained sparse approximation.
The idea is that from the converged MAP solution, we define a sparse likelihood of the Poisson
model via the replacement
\begin{equation}
L^s(\bg_s,\lambda) \doteq L(\dataset\vert \EE{P(g\vert \bg_s)}{g}, \lambda)
\end{equation}
For this sparse likelihood it is easy to compute the Laplace posterior using second derivatives. Here,
the change of variables $\rho=\ln \lambda$ will be made to ensure that $\lambda>0$. This results in an
effective log--normal density over the maximal intensity rate $\lambda$.
While we do not address hyperparameter selection for the Laplace posterior in this work, a straightforward approach, as suggested by~\citet{flaxman2017poisson}, could be to use cross validation to optimise the kernel parameters while finding the MAP estimate or to use the Laplace approximation to approximate the evidence. As in the variational case the inducing point locations $\set{\bx_l}_{l=1}^L$ will be on a regular grid over space $\X$.

Note that for the Laplace approximation, the augmentation scheme is only used to compute the MAP estimate in 
an efficient way. There are no further mean--field approximations involved. This also implies, that dependencies between $\bg_s$ and $\lambda$ are retained. 

\subsection{Predictive density} Both variational and Laplace approximation yield a posterior distribution $q$ over $\bg_s$ and $\lambda$. The GP approximation at any given points in $\X$ is given by
\begin{equation}
q(g(\bx)) = \int \int p(g(\bx) \vert \bg_s)q(\bg_s,\lambda)\;d\bg_s d\lambda,
\end{equation}
which for both methods results in a normal density. To find the posterior mean of the intensity function at a
point $\bx \in\X$ one needs to compute
\begin{equation}
\EE{Q}{\Lambda(\bx)} = \EE{Q}{\lambda \int_{-\infty}^{\infty} \sigma(g(\bx))}.
\end{equation}
For variational and Laplace posterior the expectation over $\lambda$ can be computed analytically, leaving the expectation over $g(\bx)$, which is computed numerically via quadrature methods. To evaluate the performance of inference results we are interested in computing the likelihood on test data $\dataset_{\rm test}$, generated from the ground truth. We will consider two methods:

Sampling GPs $g$ from the posterior we calculate the (log) mean of the test likelihood
\begin{equation}\label{eq:log predictive likelihood}
\begin{split}
\ell(\dataset_{\rm test}) = & \ln\EE{P}{L(\dataset_{\rm test}\vert \Lambda)\vert \dataset} \approx \ln\EE{Q}{ L(\dataset_{\rm test}\vert \Lambda)} 
\\ 
= & \ln\EE{Q}{\exp\left(-\int_{\X}\lambda\sigma(g(\bx))d\bx\right)\prod_{\bx_n\in\dataset_{\rm test}}\lambda\sigma(g(\bx_n))} 
\end{split}
\end{equation}
where the integral in the exponent is approximated by Monte--Carlo integration. The expectation is approximated by averaging over $2\times10^3$ samples from the inferred posterior $Q$ of $\lambda$ and $g$ at the observations of $\dataset_{\rm test}$ and the integration points.

\noindent
Instead of sampling one can also obtain an analytic approximation for the log test likelihood in Equation~\eqref{eq:log predictive likelihood} by a second order Taylor expansion around the mean of the obtained posterior. Applying this idea to the variational mean field posterior we get
\begin{equation}\label{eq:expandend likelihood}
\begin{split}
\ell(\dataset_{\rm test}) \approx & \ln L(\dataset_{\rm test}\vert \Lambda_Q) + \frac{1}{2}\EE{Q}{\left(\bg_s - \bmu_2^s\right)^\top \left.\mathbf{H}_{\bg_s}\right\vert_{\Lambda_Q} \left(\bg_s - \bmu_2^s\right)} \\
& + \frac{1}{2}\left.H_\lambda\right\vert_{\Lambda_Q}{\rm Var}_{Q}(\lambda),
\end{split}
\end{equation}
where $\Lambda_Q(\bx) = \EE{Q}{\lambda}\sigma(\EE{Q}{g(\bx)})$ and $\left.\mathbf{H}_{\bg_s}\right\vert_{\Lambda_Q},\ \left.H_\lambda\right\vert_{\Lambda_Q}$ are the second order derivative of the likelihood in Equation~\eqref{eq:sigmoid poisson likelihood} with respect to $\bg_s$ and $\lambda$ at $\Lambda_Q$. While an approximation only involving the first term would neglect the uncertainties in the posterior (as done by~\citet{john2018large}), the second and third term take these into account.

\section{Results}\label{sec:results}
\paragraph{Generating data from the model} To evaluate the two newly developed algorithms we generate data according to the sigmoidal Gaussian Cox process model
\begin{eqnarray}
g \sim & \bs{p}_{\rm GP}(\cdot\vert 0, k), & \\
\dataset \sim & \bs{p}_{\Lambda}(\cdot), &
\end{eqnarray}
where $\bs{p}_{\Lambda}(\cdot)$ is the Poisson process density over sets of point with $\Lambda(\bx)=\lambda\sigma(g(\bx))$ and $\bs{p}_{\rm GP}(\cdot\vert 0, k)$ is a GP density with mean $0$ and covariance function $k$. As kernel we choose a squared exponential function
\begin{equation}
k(\bx,\bx^\prime) = \theta \prod_{i=1}^d \exp\left(-\frac{(x_i - x_i^\prime)^2}{2\nu_i^2}\right),
\end{equation}
where the hyperparameters are scalar $\theta$ and length scales $\bnu=(\nu_1,\ldots,\nu_d)^\top$. Sampling of the inhomogeneous Poisson process is done via {\it thinning} \citep{lewis1979simulation,adams2009tractable}. We assume that hyperparameters are known for subsequent experiments with data sampled from the generative model. 

\paragraph{Benchmarks for sigmoidal Gaussian Cox process inference} We compare the proposed algorithms to two alternative inference methods for the sigmoidal Gaussian Cox process model. As an exact inference method we use the sampling approach of \citet{adams2009tractable}\footnote{To increase efficiency, the GP values $g$ are sampled by elliptical slice sampling~\citep{murray2010elliptical}.}. In terms of speed, a competitor is a different variational approach given by \citet{hensman2015mcmc} who proposed to discretise space $\X$ in several regular bins with size $\Delta$. Then the likelihood in Equation~\eqref{eq:sigmoid poisson likelihood} is approximated by
\begin{equation}
L(\dataset\vert \lambda\sigma(g(\bx))) \approx \prod_{i}\Po(n_i\vert \lambda\sigma(g(\bx_i))\Delta),
\end{equation}
where $\Po$ is the Poisson distribution conditioned on the mean parameter, $\bx_i$ is the centre of bin $i$, and $n_i$ the number of observations within this bin. 
Using a (sparse) Gaussian variational approximation the corresponding Kullback--Leibler divergence is minimised by gradient ascent to find the optimal posterior over the GP $g$ and a point estimate for $\lambda$. This method was originally proposed for the $\log$ Cox-process ($\Lambda(\bx)=e^{g(\bx)}$), but with the elegant GPflow package \citep{GPflow2017} implementation of the scaled sigmoid link function is straightforward. It should be noted, that this method requires numerical integration over the sigmoid link function to evaluate the variational lower bound at every spatial bin and every gradient step, since it does not make use of our augmentation scheme (see Section~\ref{sec:discussion} for discussion, how the proposed augmentation can be used for this model). We refer to this inference algorithm as `variational Gauss'. To have fair comparison between the different methods, the inducing points for all algorithms (except for the sampler) are equal and the number of bins used to discretise the domain $\X$ for the variational Gauss algorithm is set equal to the number of integration points used for the MC integration in the variational mean field and the Laplace method.

\begin{figure}
\centering
\includegraphics[width=.9\textwidth]{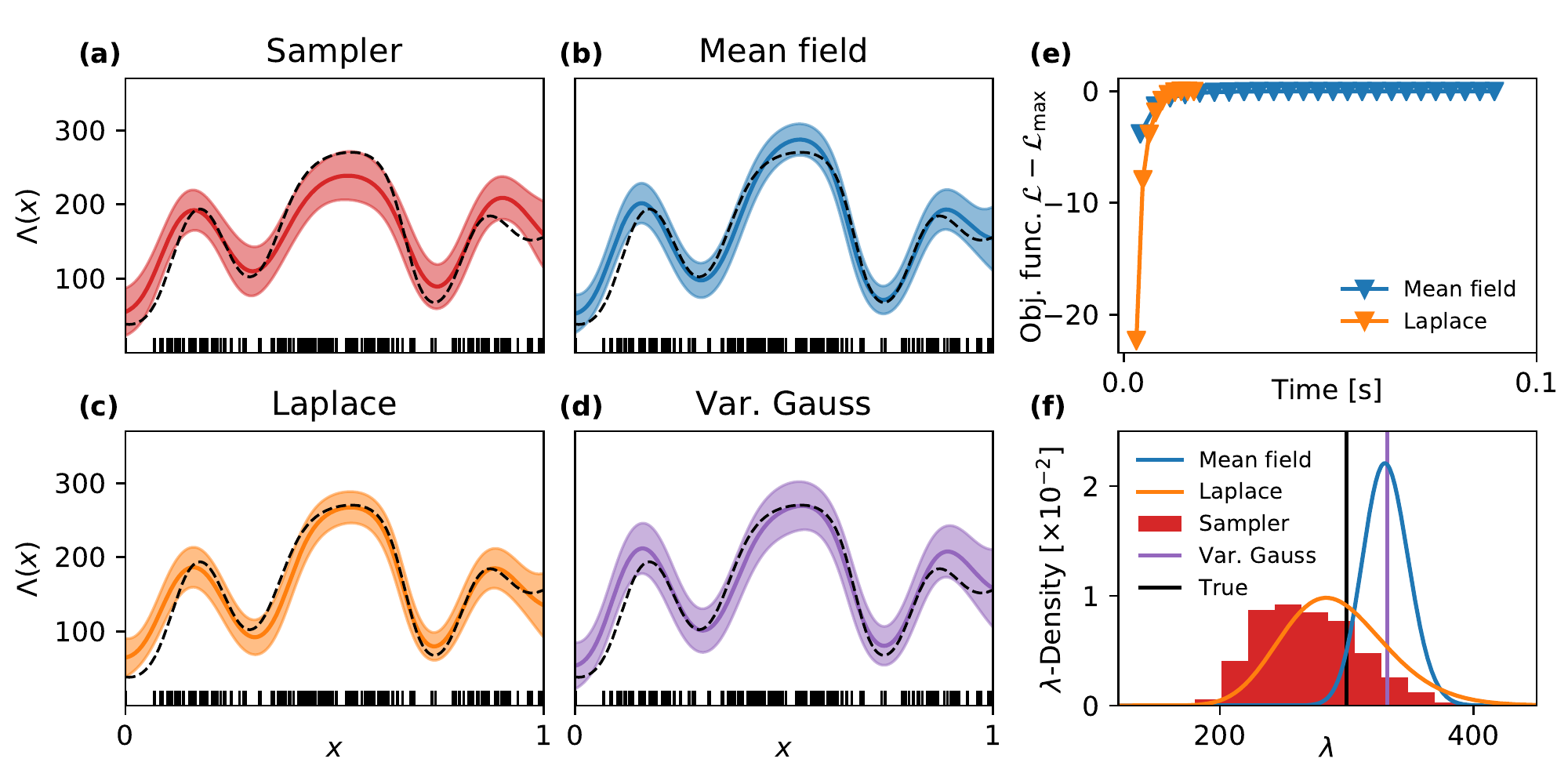}
\caption{{\bf Inference on 1D dataset.} {\bf (a)}--{\bf (d)} Inference result for sampler, mean field algorithm, Laplace approximation, and variational Gauss. Solid coloured lines denote the mean intensity function, shaded areas mean $\pm$ standard deviation, and dashed black lines the true rate functions. Vertical bars are observations $\dataset$. {\bf (e)} Convergence of mean field and EM algorithm. Objective functions (Lower bound for mean--field and log likelihood for EM algorithm, shifted such that convergence is at $0$) as function of run time (triangle marks one finished iteration of the respective algorithm). {\bf (f)} Inferred posterior densities over the maximal intensity $\lambda$. Variational Gauss provides only a point estimate. Black vertical bar denotes the true $\lambda$.}
\label{fig:fig1}
\end{figure}

\paragraph{Experiments on data from generative model} As an illustrative example we sample a one dimensional Poisson process with the generative model and perform inference with the sampler ($2\times10^3$ samples after $10^3$ burn-in iterations), the mean field algorithm, the Laplace approximation and the variational Gauss. In Figure~\ref{fig:fig1} {\bf (a)}--{\bf (d)} the different posterior mean intensity functions with their standard deviations are shown. For {\bf (b)}--{\bf (d)} $50$ regularly spaced inducing points are used. For {\bf (b)}--{\bf (c)} $2\times 10^3$ random integration points are drawn uniformly over the space $\X$, while for {\bf (d)} $\X$ is discretised into the same number of bins. All algorithms recover the true intensity well. The mean field and the Laplace algorithm show smaller posterior variance compared to the sampler. The fastest inference result is obtained by the Laplace algorithm in $0.02\ \rm{s}$, followed by the mean field ($0.09$), variational Gauss ($80$) and the sampler ($1.8\times10^3$). The fast convergence of the Laplace and the variational mean field algorithm is illustrated in Figure~\ref{fig:fig1} {\bf (e)}, where objective functions of our two algorithms (minus the maximum they converged to) is shown as a function of run time. Both algorithms reach a plateau in only a few ($\sim 6$) iterations. To compare performance in terms of log expected test likelihood $\ell_{\rm test}$ (test sets $\dataset_{\rm test}$ sampled from the ground truth), we averaged results over ten independent datasets. The posterior of the sampler yields the highest value with $875.5$, while variational ($\ell_{\rm test}=686.2$, approximation by Equation~\eqref{eq:expandend likelihood} yields $686.5$), variational Gauss ($686.7$) and Laplace ($686.1$) yield all similar results (see also Figure~\ref{fig:fig4} {\bf (a)}). The posterior density of the maximal intensity $\lambda$ is shown in Figure~\ref{fig:fig1} {\bf (f)}.

\begin{figure}
\centering
\includegraphics[width=.9\textwidth]{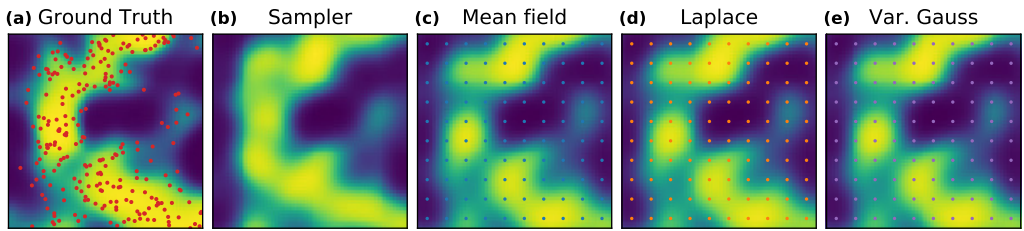}
\caption{{\bf Inference on 2D dataset.} {\bf (a)} Ground truth intensity function $\Lambda(\bx)$ with observed dataset $\dataset$ (red dots).{\bf (b)}--{\bf (e)} Mean posterior intensity of the sampler, mean field algorithm, Laplace, and variational Gauss are shown. $100$ inducing points on a regular grid (shown as coloured points) and $2500$ integration points/bins are used.}
\label{fig:fig2}
\end{figure}

In Figure~\ref{fig:fig2} we show inference results for a two dimensional Cox process example. $10\times10$ inducing points and $2500$ integration points/bins are used for mean field, Laplace and variational Gauss algorithm. The posterior mean of sampler {\bf (b)}, of the mean field {\bf (c)}, of the Laplace {\bf (d)} and of the variational Gauss algorithm {\bf (e)} recover the true intensity rate $\Lambda(\bx)$ {\bf (a)} well.

\begin{figure}
\centering
\includegraphics[width=.9\textwidth]{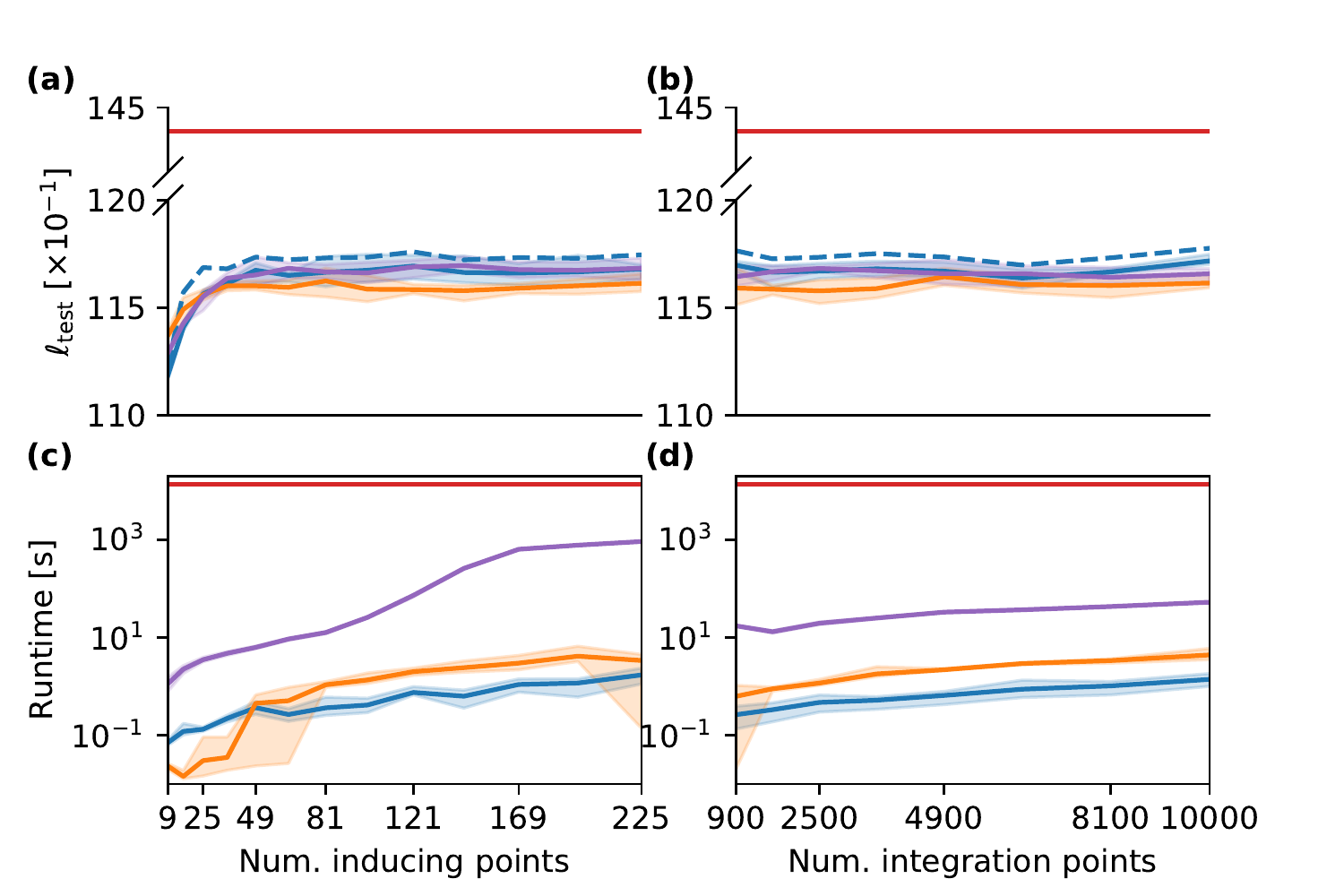}
\caption{{\bf Evaluation of inference.} {\bf (a)} The log expected predictive likelihood averaged over ten test sets as a function of the number of inducing points. Number of integration points/bins is fixed to $2500$. Results for sampler in (red), mean field (blue), Laplace (orange), and variational Gauss (purple) algorithm. Solid line denotes mean over five fits (same data), and shaded area denotes min. and max. result. Dashed blue line shows the approximated log expected predictive likelihood for the mean field algorithm. {\bf (b)} Same as (a), but as function of number of integration points. Number of inducing points is fixed to $10\times10$. Below: Run time of the different algorithms as function of number of inducing points {\bf (c)} and number of integration points {\bf (d)}. Data are the same as in Figure~\ref{fig:fig2}.}
\label{fig:fig3}
\end{figure}

\begin{figure}
\centering
\includegraphics[width=.9\textwidth]{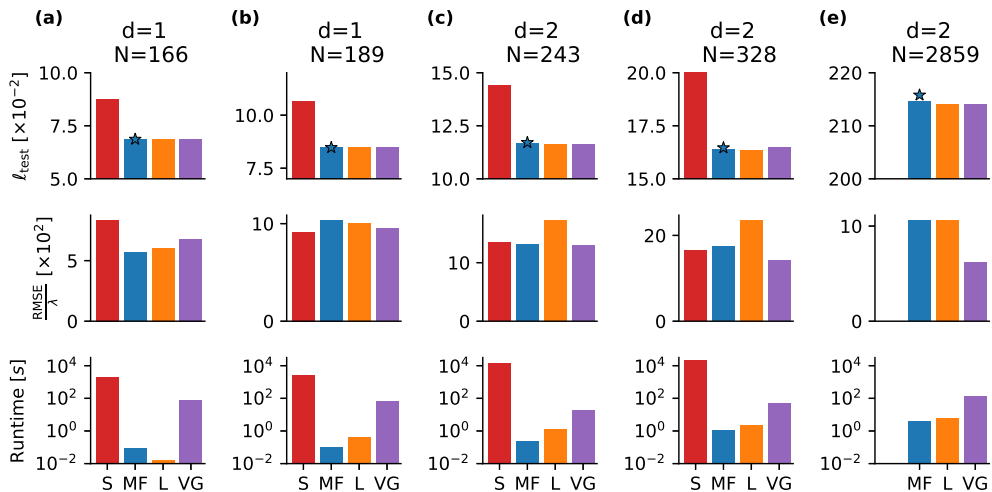}
\caption{{\bf Performance on different artificial datasets.} The sampler (S), the mean field algorithm (MF), the Laplace (L), and variational Gauss (VG) are compared on five different datasets with $d$--dimensions and $N$ observations (one column corresponds to one dataset). Top row: Log expected test likelihood of the different inference results. The star denotes the approximated test likelihood of the variational algorithm. Center row: The approximated root mean squared error (normalised by true maximal intensity rate $\lambda$). Bottom row: Run time in seconds. The dataset {\bf (e)} is intractable for the sampler due to the many observations. Data in Figure~\ref{fig:fig1} and~\ref{fig:fig2} correspond to {\bf (a)} and {\bf (c)}.}\label{fig:fig4}
\end{figure}

To evaluate the role of the number of inducing points and number of integration points we generate $10$ test sets $\dataset_{\rm test}$ from a process with the same intensity as in Figure~\ref{fig:fig2}{\bf (a)}. We evaluate the log expected likelihood (Equation~\eqref{eq:log predictive likelihood}) on these test sets and compute the average. The result is shown for different numbers of inducing points (Figure~\ref{fig:fig3}{\bf (a)} with $2500$ integration points) and different numbers of integration points (Figure~\ref{fig:fig3}{\bf (b)} with $10\times10$ inducing points). To account for randomness of integration points the fitting is repeated five times and the shaded area is between the minimum and maximum obtained by these fits. For all approximate algorithms the log predictive test likelihood saturates already for few inducing points ($\approx49\; (7\times 7)$) of the sparse GP. However, as expected, the inference approximations are slightly inferior to the sampler. The log expected test likelihood is hardly affected by the number of integration points as shown in Figure~\ref{fig:fig3} {\bf (b)}. Also the approximated test likelihood for the mean field algorithm in Equation~\eqref{eq:expandend likelihood} yields good estimates of the sampled value (dashed line in {\bf (a)} and {\bf (b)}).
In terms of runtime (Figure~\ref{fig:fig4} {\bf (c)}--{\bf (d)}) the mean field algorithm and the Laplace approximation are superior by more than one order of magnitude to the variational Gauss algorithm for this particular example. Difference increases with increasing number of inducing points.

In Figure~\ref{fig:fig4} the four algorithms are compared on five different datasets sampled from the generative model. As we observed for the previous examples the three different approximating algorithms yield qualitatively similar performance in terms of log test likelihood $\ell_{\rm test}$, but the sampler is superior. Again the approximated test likelihood in Equation~\eqref{eq:expandend likelihood} (blue star) provides good estimate of the sampled value. In addition we provide the approximated root mean squared error (RMSE, evaluated on a fine grid and normalised by maximal intensity $\lambda$) between inferred mean and ground truth. In terms of run time the mean field and Laplace algorithm are by at least on order of magnitude faster than the variational Gauss algorithm. In general, the mean--field algorithm seems to be slightly faster than the Laplace.

\begin{figure}
\centering
\includegraphics[width=\textwidth]{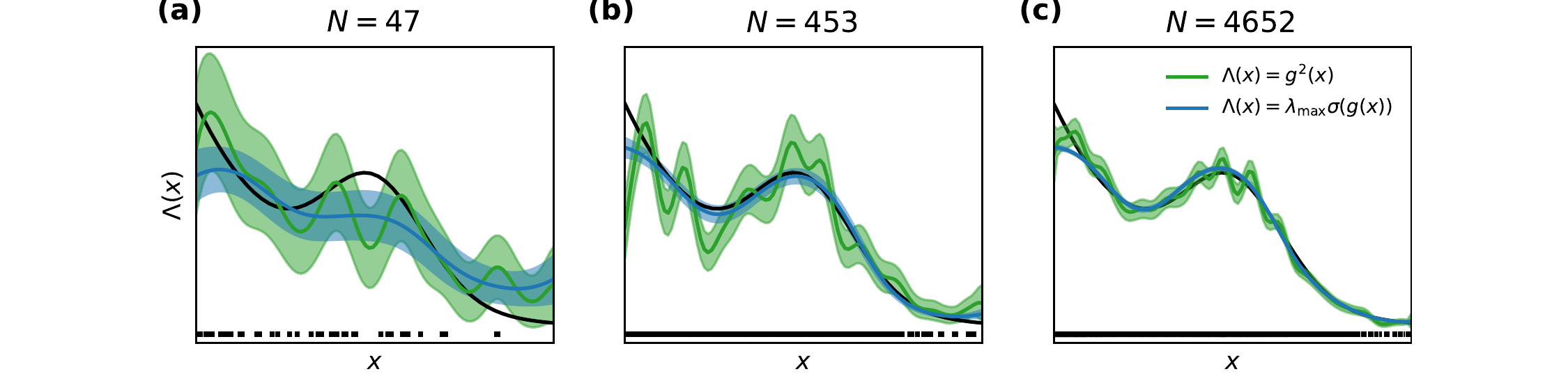}
\caption{{\bf 1D example.} Observations (black bars) are sampled from the same function (black line) scaled by {\bf (a)} 1, {\bf (b)} 10, and {\bf (c)} 100. Blue and green line show the mean posterior of the sigmoidal and squared Gaussian Cox process, respectively. Shaded area denotes mean $\pm$ standard deviation.}\label{fig:fig5}
\end{figure}

\paragraph{General datasets and comparison to the approach of Lloyd et al.} Next, we test our variational mean field algorithm on datasets not coming from the generative model. On such datasets we do not know, whether our model provides a good prior. As discussed previously an alternative model was proposed by~\citet{lloyd2015variational} making use of the link function $\Lambda(\bx)=g^2(\bx)$. While the sigmoidal Gaussian Cox process with the proposed augmentation scheme has analytic updates for the variational posterior, in case of the squared Gaussian Cox process the likelihood integral can be solved analytically and does not need to be sampled (if the kernel is a squared exponential and the domain is rectangular). Both algorithms rely on the sparse GP approximation. To compare the two methods empirically first we consider one dimensional data generated using a known intensity function. We choose $\Lambda(x)=2\exp(-x/15) + \exp(-(x-25)^2/100)$ on an interval $[0,50]$ already proposed by~\citet{adams2009tractable}. We generate three training and test sets, where we scale this rate function by factors of $1,\;10,$ and $100$ and fit the sigmoidal and squared Gaussian Cox process with their corresponding variational algorithm to each training set\footnote{We thank Chris Lloyd and Tom Gunter for providing the code for inferring the variational posterior of the squared Gaussian Cox process.}. The number of inducing points is $40$ in this example. For our variational mean field algorithm we used $5000$ integration points. The posterior intensity $\Lambda(\bx)$ for the three datasets can be seen in Figure~\ref{fig:fig5}. The model with the sigmoidal link function infers smoother posterior functions with smaller variance compared to the posterior with the squared link function. For datasets shown in Figure~\ref{fig:fig5} we run the fits five times and report mean and standard deviation of runtime, RMSE and log expected test likelihood $\ell_{\rm test}$ in Table~\ref{tab:table1}. Run times of the two algorithms are comparable, where for the intermediate dataset the algorithm with the squared link function is faster while for the largest data set the one with the sigmoidal link function converges first. RMSE and $\ell_{\rm test}$ are also comparable except for the intermediate dataset, where the sigmoidal model is the superior one.

\begin{table}[]
\begin{tabular}{l|lll|lll}
         & \multicolumn{3}{c|}{$\Lambda(x) = \lambda_{\rm max}\sigma(g(x))$} & \multicolumn{3}{c}{$\Lambda(x) = g^2(x)$}   \\ \hline
 $N$        & Runtime {[}s{]}     & RMSE              & $\ell_{\rm test}$      & Runtime {[}s{]} & RMSE & $\ell_{\rm test}$  \\ \hline
$47$   & $0.27\pm0.30$       & $0.24\pm 0.02$    & $-43.43 \pm 0.42$      & $0.41\pm 0.05$  &  $0.24$   & $-44.26 \pm 0.09$  \\
$453$  & $0.50\pm 0.04$      & $0.97\pm 0.13$    & $720.81 \pm 0.28$      & $0.23\pm 0.05$  &   $2.11$   & $710.43\pm1.38$    \\
$4652$ & $0.41\pm 0.01$      & $7.68\pm0.75$     & $17497.31 \pm 2.13$    & $0.79\pm 0.09$  &  $8.16$    & $17496.75\pm 1.65$ \\
\hline
\end{tabular}
\caption{{\bf Benchmarks for Figure~\ref{fig:fig5}} The mean and standard deviation of runtime, RMSE, and log expected test likelihood for Figure~\ref{fig:fig5}{\bf (a)}--{\bf (c)} obtained from $5$ fits. Note that the RMSE for $\Lambda(\bx)=g^2(\bx)$ has no standard deviation, because the inference algorithm is deterministic.}\label{tab:table1}
\end{table}

\begin{figure}
\centering
\includegraphics[width=\textwidth]{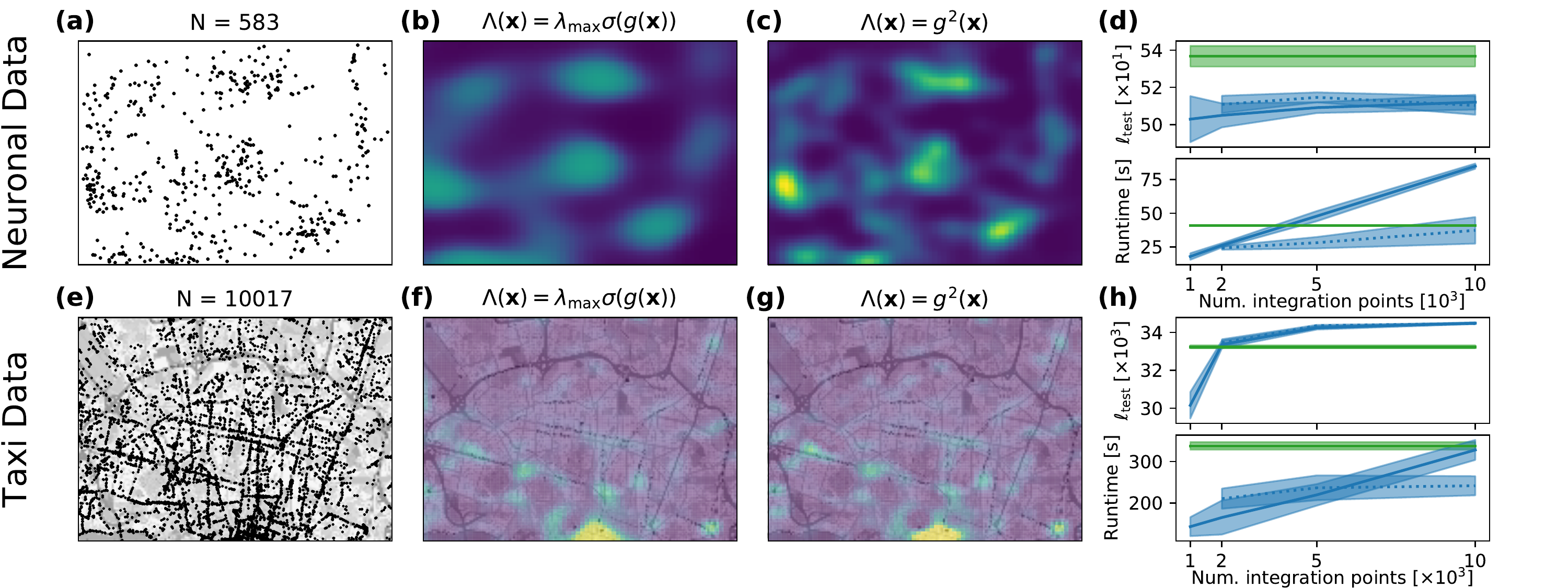}
\caption{{\bf Fits to real world datasets.} {\bf (a)} Position of the mouse while the recorded neuron spiked. {\bf (b)} Posterior mean obtained by the variational mean--field algorithm for the sigmoidal Gaussian Cox process. {\bf (c)} Same as in (b) for the variational approximation of the squared Gaussian Cox process. {\bf (d)} Log expected test--likelihood $\ell_{\rm test}$ and runtime as function of number of integration points for both algorithms. The dotted line is obtained by first fitting the sigmoidal model with $1000$ integration points and then with the number that is indicated on the x-axis. Shaded area is mean $\pm$ standard deviation obtained in $5$ repeated fits.  {\bf (e)}--{\bf (h)} Same as (a)--(d), but for a dataset, where the observations are positions of taxi pick--ups in the city of Porto.}\label{fig:fig6}
\end{figure}

Next we deal with two real world two dimensional datasets for comparison. The first one is neuronal data, where spiking activity was recorded from a mouse, that was freely moving in an arena~\citep{gridcelldata,sargolini06}. Here we consider as data $\dataset$ the position of the mouse when the recorded cell fired and the observations are randomly assigned to either training or test set. In Figure~\ref{fig:fig6} {\bf(a)} the observations in the training set ($N=583$) are shown. In Figure~\ref{fig:fig6} {\bf (b)} and {\bf (c)} the variational posterior's mean intensity $\Lambda(\bx)$ is shown obtained for the sigmoidal and the squared link function, respectively, inferred with a regular grid of $20\times 20$ inducing points. As in Figure~\ref{fig:fig5} we see that the sigmoidal posterior is the smoother one. The major difference between the two algorithms (apart from the link function) is the fact that for the sigmoidal model we are required to sample an interval over the space. We investigate the effect of the number of integration points in terms of runtime\footnote{Note, that - in contrast to Figures~\ref{fig:fig3} and~\ref{fig:fig4} - the runtime is displayed on linear scale, meaning both algorithms are of same order of magnitude.} and log expected test likelihood in Figure~\ref{fig:fig6} {\bf (d)}. First, we observe regardless of the number of integration points that the variational posterior of the squared link function yields the superior expected test likelihood. For the sigmoidal model the test likelihood does not improve significantly with more integration points. Runtimes of both algorithms are comparable, when $5000$ integration points are chosen. A speed up for our mean field algorithm is achieved by first fitting the model with $1000$ integration points and once converged, redrawing the desired number of integration points and rerun the algorithm (dotted line in Figure~\ref{fig:fig6}{\bf (d)}). This method allows for a significant speed up without loss in terms of test likelihood $\ell_{\rm test}$. The variational mean-field algorithm with the sigmoid link function is faster with up to $5000$ integration points and equally fast with $10000$ integration points.

As second dataset we consider the Porto taxi dataset~\citep{moreira2013predicting}. These data contain trajectories of taxi travels from the years $2013/14$ in the city of Porto. As~\citet{john2018large} we consider the pick-ups as observations of a Poisson process\footnote{As \citet{john2018large} report some regions to be highly peaked we consider only pickups happening within the coordinates ($41.147,-8.58$) and ($41.18,-8.65$) in order to exclude those regions.}. We consider $20000$ taxi rides randomly split into training and test set ($N=10017$ and $N=9983$, respectively). The training set is shown in Figure~\ref{fig:fig6}{\bf (e)}. Inducing points are positioned on a regular grid of $20\times 20$. The variational posterior mean of the respective intensity is shown in Figure~\ref{fig:fig6} {\bf (f)} and {\bf (g)}. With as many data points as in these data the differences between the two models are more subtle as compared to {\bf (b)} and {\bf (c)}. In terms of test likelihood $\ell_{\rm test}$ the variational posterior of the sigmoidal model (with $\geq 2000$ integration points) outperforms the model with squared link function (Figure~\ref{fig:fig6} {\bf (h)}). For similar test likelihoods $\ell_{\rm test}$ our variational algorithm is $\sim 2\times$ faster than the variational posterior with squared link function. The results show that the choice of number of integration points reduces to the question of speed vs accuracy trade--off. As for the previous dataset, the strategy of first fitting the posterior with $1000$ integration points and then with the desired number of integration points (dotted line) proves that we can get a significant speed up without loosing predictive power.

\section{Discussion and Outlook}\label{sec:discussion}
Using a combination of two known variable augmentation methods, we 
derive a conjugate representation for the posterior measure of a sigmoidal Gaussian Cox process.
The approximation of the augmented posterior by a simple mean field factorisation yields an efficient 
variational algorithm. The rationale behind this method is that the variational updates in the conjugate
model are explicit and analytical and do not require (black--box) gradient descent methods.
In fact, a comparison with a different variational algorithm for the same model - not based on augmentation, but on 
direct approximation of the posterior with a Gaussian - shows that the qualities of inference for both 
approaches are similar, while the mean field algorithm is at least one order of magnitude faster.
We use the same variable augmentation method for computation of the MAP estimate 
for the (unaugmented) posterior by a fast EM algorithm. This is finally applied to the calculation
of Laplace's approximation. Both methods yield an explicit result for the approximate 
GP posterior. Since the corresponding effective likelihood contains a continuum of the GP 
latent variables, the exact computations of means and marginal variances would require 
the inversion of a linear operator instead of a simpler matrix inverse. While for specific priors, 
this problem could be solved by PDE or ODE methods, we resort to a well known sparse GP approach with inducing points in this paper. We can apply this to arbitrary kernels but need to solve spatial integrals over the domain. These can be (at least for moderate dimensionality) well approximated by simple Monte Carlo integration. Advantage of this approach is, that one is not limited to rectangular domains. The only requirement is that the volume $|\X|$ is known. An alternative Poisson model for which similar spatial integrals can be performed analytically \citep{lloyd2015variational} within the sparse GP approximation (limited to squared exponential kernels and rectangular domains) is based on a quadratic link function
\citep{lloyd2015variational,flaxman2017poisson,john2018large}. We compare our variational algorithm with the variational algorithm of~\citet{lloyd2015variational} on different datasets and observe that both algorithms act on the same order of magnitude in terms of runtime (with slight advantages for our variational mean field algorithm). As expected, we show that whether one or the other model is better in predictive power is highly data dependent.

As an alternative to the Monte Carlo integration in our approach we could avoid the infinite dimensionality of the latent GP from the beginning by working with a binning scheme for the Poisson observations as in~\citet{hensman2015mcmc}.
It would be straightforward to adopt our augmentation method to this case. The resulting Poisson likelihoods would then be augmented by pairs of Poisson and P\'olya--Gamma variables (see~\citet{donner2017inverse}) for each bin. This approach could be favourable when the number of observed data points becomes very large, because the discretisation method does not scale with the number data points but with the resolution of discretisation. However, we do expect, that any approach based on either spatial discretisation or on the sparse, inducing point method would become problematic for large or high dimensional domains $\X$.  Alternative methods based on spectral representations of kernels~\citep{knollmuller2017inference,john2018large} are promising 
for tackling those problems.

It will be interesting to apply the variable augmentation method to other Bayesian models with the sigmoid link function. For example, the inherent boundedness of the resulting intensity can be crucial for point processes such as the nonlinear {\it Hawkes process}~\citep{hawkes1971spectra} which is widely used for modelling stock market data~\citep{embrechts2011multivariate} or seismic activity~\citep{ogata1998space}. For other point process models the sigmoid function appears naturally. We mention the kinetic Ising model, a Markov jump process~\citep{donner2017inverse} which was originally introduced to model the dynamics of classical spin systems in physics. More recently it was used to model the joint activity of neurons~\citep{dunn2015correlations}. Finally, a Gaussian process density model introduced by~\citep{murray2009gaussian} can be treated by the augmentations developed in this work~\citep{donner2018efficient}. 

\acks{CD was supported by the Deutsche Forschungsgemeinschaft (GRK1589/2) and partially funded by Deutsche Forschungsgemeinschaft (DFG) through grant CRC 1294 ``Data Assimilation'', Project (A06) ``Approximative Bayesian inference and model selection for stochastic differential equations (SDEs)''.}


\newpage

\appendix
\appendix

\section{Poisson processes}\label{app:poisson process}
In this paragraph we briefly summarise those properties of a Poisson process, which are relevant for this work.
For a thorough and more complete description we recommend the concise book by \citet{kingman1993poisson}, particularly chapter 3 and 5.

We consider a general space $\calZ$ and a countable subset $\Pi_\calZ = \set{\bz;\bz\in \mathcal{Z}}$.

\paragraph{Definition of a Poisson process} A random countable subset $\Pi_\calZ \subset \calZ$ is a Poisson process on $\calZ$, if 
\begin{enumerate}[i)]
\item for any sequence of disjoint subsets $\set{\calZ_k\subset \calZ}_{k=1}^K$ the cardinality of the union 

$N(\calZ_k)\doteq \vert \set{\Pi_\calZ\cap\calZ_k}\vert$ is independent of $N(\calZ_l)$ for all $l\neq k$.
\item  $N(\calZ_k)$ is Poisson distributed with mean $\int_{\calZ_k}\Lambda(\bz)d\bz$, and mean measure $\Lambda(\bz):\X \rightarrow \R^{+}$.
\end{enumerate}

If the mean measure is constant ($\Lambda(\bz)=\Lambda$) the Poisson process is {\it homogeneous}, and {\it inhomogeneous} otherwise.

\paragraph{Campbell's Theorem} Let $\Pi_\calZ$ be a Poisson process on $\calZ$ with mean measure $\Lambda(\bz)$. Furthermore, we define a function $h(\bz):\calZ\rightarrow \R$ and the sum 
\begin{equation}
H(\Pi_\calZ) = \sum_{\bz\in\Pi_\calZ}h(\bz).
\end{equation}
If $\Lambda(\bz)<\infty$ for $\bz~\in\calZ$, then
\begin{equation}\label{eq:characteristic functional}
\EE{P_{\Lambda}}{e^{\xi H(\Pi_\calZ)}} = \exp\left\{\int_{\calZ} \left(e^{\xi h(\bz)} - 1\right)\Lambda(\bz)d\bz \right\},
\end{equation}
for any $\xi\in\mathbb{C}$, such that the integral converges. $P_{\Lambda}$ is the probability measure of a Poisson process with intensity $\Lambda(\bz)$. Mean and variance are obtained as 
\begin{align}
\EE{P_{\Lambda}}{H(\Pi_\calZ)} = & \int_{\calZ}h(\bz)\Lambda(\bz)d\bz, \\
{\rm Var}_{P_{\Lambda}}\left[H(\Pi_\calZ)\right] = & \int_{\calZ}[h(\bz)]^2\Lambda(\bz)d\bz.
\end{align}
Note, that Equation~\eqref{eq:characteristic functional} defines the {\it characteristic functional} of a Poisson process.

\paragraph{Marked Poisson process} Let $\Pi_\calZ=\set{\bz_n}_{n=1}^N$ a Poisson process on $\calZ$ with intensity  $\Lambda(\bz)$. Then $\Pi_{\hat{\calZ}}=\set{(\bz_n,\bs{m}_{n})}_{n=1}^N$ is again a Poisson process on the product space $\hat{\calZ} = \calZ\times\calM$, if $\bs{m}_n\sim p(\bs{m}_n\vert \bz_n)$ is drawn independently at each $\bz_n$. The $\bs{m}_n\in \calM$ are the so--called `marks', and the resulting Process is a {\it marked Poisson process} with intensity
\begin{equation}
\Lambda(\bz,\bs{m}) = \Lambda(\bz)p(\bs{m}\vert \bz).
\end{equation}
It is straightforward to extend Campbell's theorem and to show that the characteristic functional of such a process is 
\begin{equation}\label{eq:marked characteristic functional}
\EE{P_{\Lambda}}{e^{\xi H(\Pi_{\hat{\calZ}})}} = \exp\left\{\int_{\hat{\calZ}} \left(e^{\xi h(\bz,\bs{m})} - 1\right)\Lambda(\bz,\bs{m})\;d\bs{m}d\bz \right\},
\end{equation}
with $h(\bz,\bs{m}):\hat{\calZ}\rightarrow\R$ and $H(\Pi_{\hat{\calZ}})=\sum_{(\bz,\bs{m})\in \Pi_{\hat{\calZ}}}h(\bz,\bs{m})$.

\section{The P\'olya-Gamma density}\label{app:polya gamma}
The P\'olya-Gamma density~\citep{polson2013bayesian} has the useful property, that it allows to represent the inverse hyperbolic cosine by an infinite Gaussian mixture as
\begin{equation}
\cosh^{-b}(c/2) = \int_{0}^\infty \exp\left(-\frac{c^2}{2}\omega\right)\PG(\omega\vert b,0)d\omega,
\end{equation}
with parameter $b>0$. Furthermore, one can define a {\it tilted P\'olya-Gamma density} as 
\begin{equation}
\PG(\omega\vert b, c)= \frac{\exp\left(-\frac{c^2}{2}\omega\right)}{\cosh^{-b}(c/2)}\PG(\omega\vert b,0).
\end{equation}
From those two equations the moment generating function can be obtained from the basic definition, being
\begin{equation}
\int_{0}^\infty e^{\xi\omega}\PG(\omega\vert b,c)d\omega = \frac{\cosh^b(c/2)}{\cosh^b\left(\sqrt{\frac{c^2/2-\xi}{2}}\right)},
\end{equation}
and differentiating with respect to $\xi$ at $\xi=0$ yields the first moment
\begin{equation}
\EE{\PG}{\omega} = \frac{b}{2c}\tanh\left(c/2\right).
\end{equation}

\section{Variational inference for stochastic processes}\label{app:variational inference}
\paragraph{Densities for random processes}
A stochastic process $X$  with probability measure $P(X)$ often has no 
density with respect to Lebesgue measure, since $X$ can be an infinite dimensional object
such as a function for the case of a Gaussian process. However, one can define densities 
with respect to another (reference) measure $R(X)$ written as
\begin{equation}\label{eq:radon nikodym}
\bs{p}(X) = \frac{dP}{dR}(X),
\end{equation}
if $R(X)$ is absolutely continuous with respect to $P(X)$ (if $R(X)=0$ then $P(X)=0$).  
Using such a density, expectations are
\begin{equation}
\EE{P}{f(X)} = \int f(X)dP(X) = \int f(x)\bs{p}(x)dR(X) = \EE{R}{f(x)\bs{p}(x)}.
\end{equation}
The density in Equation~\eqref{eq:radon nikodym} is known as the {\it Radon--Nikod\'ym derivative} of $R$ with respect to $P$~\citep{konstantopoulos2011radon}. 
\paragraph{Poisson process density} As specific example consider the prior density of the Poisson process in Equation~\eqref{eq:augmented likelihood}, which is defined with respect to a reference measure
\begin{equation}
\bs{p}_{\Lambda}(\Pi_\calZ) = \frac{dP_\Lambda}{dP_{\Lambda_0}}(\Pi_\calZ) = \exp\left(-\int_{\calZ}(\Lambda(\bz) - \Lambda_0(\bz))d\bz\right)\; \prod_{\bz_n \in \Pi_\calZ} \frac{\Lambda(\bz_n)}{\Lambda_0(\bz_n)},
\end{equation}
where $P_{\Lambda_0}$ is the probability measure with intensity $\Lambda_0$ and the expectation is defined as 
\begin{equation}\label{eq:poisson expectation}
\EE{P_\Lambda}{\sum_{\bz_n\in \Pi_\calZ} u(\bz_n)} = \EE{P_{\Lambda_0}}{\bs{p}_{\Lambda}(\Pi_\calZ)\; \sum_{\bz_n\in \Pi_\calZ} u(\bz_n)}.
\end{equation}
Calculating the expectation of $e^{\xi H(\Pi_\calZ)}$ with Equation~\eqref{eq:poisson expectation} we identify the characteristic function of a Poisson process (see Equation~\eqref{eq:marked characteristic functional}) with intensity $\Lambda(\bz)$.
\paragraph{Kullback-Leibler divergence} 
Using these densities we can express the Kullback-Leibler divergence between two probability measures. 

 The KL--divergence between $\bs{q}(X)$ and $\bs{p}(X)$ is defined as
\begin{equation}
{\rm D}_{\rm KL}(Q\Vert P) = \EE{Q}{\log\frac{dQ}{dP}(X)} = \int \log\frac{\bs{q}(X)}{\bs{p}(X)}dQ(X),
\end{equation}
where 
\begin{equation}
\bs{q}(X) = \frac{dQ}{dR}(X),
\end{equation}
and where $R(X)$ also is absolutely continuous to $Q(X)$.
The KL--divergence does not depend on the reference measure $R(X)$.

\section{The posterior point process is a marked Poisson process}\label{app:optimal posterior poisson}
Here we prove  that the optimal variational posterior point process in Equation~\eqref{eq:optimal marked process} again is a Poisson process using Campbell's theorem. As posterior process in Equation~\eqref{eq:optimal marked process} one gets
\begin{equation}
\bs{q}(\Pi_{\mathcal{Z}}) = \frac{dQ}{dP_\lambda}(\Pi_{\mathcal{Z}}) = \frac{\prod_{\bs{z}_m\in \Pi_{\mathcal{Z}}} e^{f(\bs{z}_m)}}{\EE{P_\lambda}{\prod_{\bs{z}_m\in \Pi_{\mathcal{Z}}} e^{f(\bs{z}_m)}}} = \frac{\prod_{\bs{z}_m\in \Pi_{\mathcal{Z}}} e^{f(\bs{z}_m)}}{\exp\left(\int_{\mathcal{Z} }(e^{f(\bs{z})} - 1) \lambda(\bs{z}) d\bs{z}\right)},
\end{equation}
where $\Pi_\mathcal{Z}$ is some random set of points on space $\mathcal{Z}$ and $P_\lambda$ is a random Poisson measure with intensity $\lambda(\bs{z})$. To proof, that the resulting point process density $\bs{q}(\Pi_{\mathcal{Z}})$ is again a Poisson process we calculate the characteristic functional for some arbitrary function $h:\mathcal{Z}\rightarrow \R$
\begin{equation}
\begin{split}
\EE{Q}{\prod_{\bs{z}_m\in \Pi_\mathcal{Z}} e^{h(\bs{z}_m)}} = & \frac{\EE{P_\lambda}{\prod_{\bs{z}_m\in \Pi_{\mathcal{Z}}} e^{h(\bs{z}_m) + f(\bs{z}_m)}}}{\exp\left(\int_{\mathcal{Z} }(e^{f(\bs{z})} - 1) \lambda(\bs{z}) d\bs{z}\right)} \\
= & \frac{\exp\left(\int_{\mathcal{Z} }(e^{h(\bs{z})+f(\bs{z})} - 1) \lambda(\bs{z}) d\bs{z}\right)}{\exp\left(\int_{\mathcal{Z} }(e^{f(\bs{z})} - 1) \lambda(\bs{z}) d\bs{z}\right)} \\
= &\exp\left(\int_{\mathcal{Z} }(e^{h(\bs{z})} - 1) e^{f(\bs{z})}\lambda(\bs{z}) d\bs{z}\right)\\
= &\exp\left(\int_{\mathcal{Z} }(e^{h(\bs{z})} - 1) \Lambda_Q(\bs{z}) d\bs{z}\right).
\end{split}
\end{equation}
We identify the last row as the generating functional of a Poisson process~\eqref{eq:marked characteristic functional} with $\xi=1$. The intensity of the process is $\Lambda_Q(\bs{z})=e^{f(\bs{z})}\lambda(\bs{z})$. With the fact that a Poisson process is uniquely characterised by its generating function~\citep[chap. 3]{kingman1993poisson}, the proof is complete.

\section{Sparse Gaussian process approximation}\label{app:sparse GP}
To solve the inference problem for the function $g$, we define a sparse GP, using the same 
prior $P$, but by an effective likelihood which depends on a finite set of function values $\bg_s = (g_1,\ldots,g_L)^\top$ 
only. Hence, we get 
\begin{equation}\label{eq:app sparse gp}
\frac{dQ_2^s}{dP}(g) = \bs{q}^s_2(\bg_s) 
\end{equation}
and the sparse posterior measure is
\begin{equation}\label{eq:app posterior measure}
dQ^s_2(g) = \bs{q}^s_2(\bg_s)dP(g)= dP(g\vert\bg_s)\times \bs{q}^s_2(\bg_s)dP(\bg_s),
\end{equation}
where the last equality holds true, since Equation~\eqref{eq:app sparse gp} only depends on $\bg_s$.
The KL--divergence between the full posterior density
\begin{equation}
\bs{q}_2(g) = \frac{dQ_2}{dP}(g) = \frac{e^{U(g)}}{\EE{P}{e^{U(g)}}}
\end{equation}
and the sparse one $\bs{q}_2^s(\bg_s)$ is given by
\begin{equation}\label{eq:app sparse kl}
\begin{split}
{\rm D}_{\rm KL}(Q_2^s\Vert Q_2) & = \EE{Q_2^s}{\log \frac{\bs{q}_2^s(\bg_s)}{\bs{q}_2(g)}} = \EE{P(\bg_s)}{\bs{q}_2^s(\bg_s) \EE{P(g\vert\bg_s)}{\log \frac{\bs{q}_2^s(\bg_s)}{e^{U(g)}}}} + {\rm const.} \\
& = \EE{P(\bg_s)}{\bs{q}_2^s(\bg_s) \log \frac{\bs{q}_2^s(\bg_s)}{e^{\EE{P(g\vert\bg_s)}{U(g)}}}} + {\rm const.}
\end{split}
\end{equation}
From this we derive directly the posterior density for the sparse GP
\begin{equation}
\bs{q}_2^s(g) \propto e^{U^s(\bg_s)},
\end{equation}
with the sparse log--likelihood
\begin{equation}\label{eq:U function app}
U^s(\bg_s) = \EE{P(g\vert \bg_s)}{U(g)} = \int U(g)dP(g\vert\bg_s).
\end{equation}

\section{Lower bound \& hyperparameter optimization}\label{app:lower bound}
The lower bound in Equation~\eqref{eq:lower bound} is given by
\begin{equation}
\begin{split}
\mathcal{L}(\bs{q}) = & \EE{Q}{\log \frac{L(\dataset, \bomega_N, \Pi_{\hat{\X}}\vert g, \lambda )}{\bs{q}_1(\bomega_N)\bs{q}_1(\Pi_{\hat{\X}})\bs{q}_2^s(g)\bs{q}_2(\lambda)}} \\
= & \int_{\hat{\X}}  \left(\EE{Q}{f(\omega,-g(\bx))} - \EE{Q}{\log \Lambda_{1}} + \EE{Q}{\log \lambda} + 1\right) \Lambda_{1}(\bx,\omega)d\bx d\omega \\
& - \int_{\hat{\X}}\Lambda_{1}(\bx,\omega)d\bx d\omega \\
& + \sum_{n=1}^N\left(\EE{Q}{f(\omega_n,g_n)} + \EE{Q}{\log\lambda} - \cosh\left(\frac{c_1^{(n)}}{2}\right) + \frac{\left(c_1^{(n)}\right)^2}{2}\EE{Q}{\omega_n} \right)
\\
& -\frac{1}{2}trace(K_s^{-1}(\Sigma_2^s + \boldsymbol{\mu}_2^s(\boldsymbol{\mu}_2^s)^\top))- \frac{1}{2}\log \det(2\pi K_s) + \frac{1}{2}\log\det(2\pi e \Sigma_2^s) \\
& + \alpha_0\log \beta_0 - \log(\Gamma(\alpha_0)) + (\alpha_0 - 1)\EE{Q}{\log \lambda} - \beta_0 \EE{Q}{\lambda} \\
& + \alpha_{2} - \log \beta_{2} + \log \Gamma(\alpha_{2}) + (1-\alpha_{2})\psi(\alpha_{2}).
\end{split}
\end{equation}
To optimise the covariance kernel parameters $\boldsymbol{\theta}$ we differentiate the lower bound with respect to these parameters and perform then gradient ascent. The gradient for one specific parameter $\theta$ is given by
\begin{equation}
\begin{split}
\frac{\partial \mathcal{L}(\bs{q})}{\partial \theta} = & \int_{\hat{\X}}  \frac{\partial \EE{Q}{f(\omega,-g(\bx))}}{\partial \theta}\Lambda_{1}(\bx,\omega)d\bx d\omega + \sum_{n=1}^N\frac{\partial \EE{Q}{ f(\omega_n,g(\bx_n))}}{\partial \theta}
\\
& -\frac{1}{2}\frac{trace(K_s^{-1}(\Sigma^s_2 + \boldsymbol{\mu}^s_2(\boldsymbol{\mu}_2^s)^\top))}{\partial \theta}- \frac{1}{2}\frac{\partial \log \det(2\pi K_s)}{\partial \theta} \\
=  & \int_{\hat{\X}}  \frac{\partial \EE{Q}{f(\omega,-g(\bx))}}{\partial \theta}\Lambda_{1}(\bx,\omega)d\bx d\omega + \sum_{n=1}^N\frac{\partial \EE{Q}{f(\omega_n,g(\bx_n))}}{\partial \theta} \\
& +\frac{1}{2}trace\left(K_s^{-1}\frac{\partial K_s}{\partial \theta} K_s^{-1}(\Sigma^s_2 + \boldsymbol{\mu}^s_2(\boldsymbol{\mu}^s_2)^\top)\right)\\
& - \frac{1}{2}trace\left(K_s^{-1}\frac{\partial K_s}{\partial \theta}\right).
\end{split}
\end{equation}
The derivatives of function $\EE{Q}{f(\omega,g(\bx))}$ are
\begin{equation}
\begin{split}
\frac{\partial \EE{Q}{f(\omega,g(\bx))}}{\partial \theta} = & \frac{1}{2}\left(\frac{\partial \EE{Q}{g(\bx)}}{\partial \theta} - \frac{\partial \EE{Q}{g(\bx)^2}}{\partial \theta}\EE{Q}{\omega}\right),
\end{split}
\end{equation}
with
\begin{equation}
\begin{split}
\frac{\partial \EE{Q}{ g(\bx)}}{\partial \theta} = & \frac{\partial \boldsymbol\kappa(\bx)}{\partial \theta}\boldsymbol{\mu}_2^s, \\
\frac{\partial \EE{Q}{g(\bx)^2}}{\partial \theta} = & \frac{\partial \tilde{k}(\bx,\bx)}{\partial \theta} + \frac{\partial \boldsymbol{\kappa}(\bx)}{\partial \theta}^\top \left(\Sigma^s_2 + \boldsymbol{\mu}^s_2(\boldsymbol{\mu}^s_2)^\top\right) \boldsymbol{\kappa}(\bx) + \boldsymbol{\kappa}(\bx)^\top \left(\Sigma_2^s + \boldsymbol{\mu}_2^s(\boldsymbol{\mu}_2^s)^\top\right) \frac{\partial \boldsymbol{\kappa}(\bx)}{\partial \theta},
\end{split}
\end{equation}
where $\boldsymbol{\kappa}(\bx)=\boldsymbol{k}_s(\bx)^\top K_s^{-1}$ and $\tilde{k}(\bx,\bx)=k(\bx,\bx) - \bs{k}_s(\bx)K_s^{-1}\bs{k}_s(\bx)^\top$. The remaining two terms are:
\begin{equation}
\begin{split}
\frac{\partial \tilde{k}(\bx,\bx)}{\partial \theta} = & \frac{\partial k(\bx,\bx)}{\partial \theta} - \frac{\partial \boldsymbol{\kappa}(\bx)}{\partial \theta}\boldsymbol{k}_s(\bx) - \boldsymbol{\kappa}(\bx)\frac{\partial \boldsymbol{k}_s(\bx)}{\partial \theta}, \\
\frac{\partial \boldsymbol{\kappa}(\bx)}{\partial \theta} = & \frac{\partial \boldsymbol{k}_s(\bx)^\top}{\partial \theta}K_s^{-1} - \boldsymbol{k}_s(\bx)K_s^{-1}\frac{\partial K_s}{\partial \theta}K_s^{-1}.
\end{split}
\end{equation}
After each variational step the hyperparameters are updated by
\begin{equation}
\boldsymbol{\theta}_{\rm new} = \boldsymbol{\theta}_{\rm old} + \varepsilon\frac{\partial \mathcal{L}(q)}{\partial \boldsymbol{\theta}},
\end{equation}
where $\varepsilon$ is the step size.

\vskip 0.2in
\bibliography{bib}

\end{document}